\documentclass[a4paper,fleqn]{cas-sc}

\usepackage[authoryear]{natbib}
\usepackage{amssymb}
\usepackage{lipsum}
\usepackage{stfloats}
\usepackage{tabularx}
\usepackage{booktabs}
\usepackage{array} 
\usepackage{amsmath}
\usepackage{algorithm}
\usepackage{algpseudocode}
\usepackage[nameinlink]{cleveref}
\usepackage{caption}
\usepackage{multirow}
\usepackage{subcaption}
\usepackage{hyperref}
\usepackage{graphicx}
\usepackage{float}
\begin{document}
\let\WriteBookmarks\relax
\def\floatpagepagefraction{1}
\def\textpagefraction{.001}

\shorttitle{}

\shortauthors{Patade et~al.}

\title [mode = title]{Optimizing Path Planning using Deep Reinforcement Learning for UGVs in Precision Agriculture}                      
% \tnotemark[1,2]

% \tnotetext[1]{This document is the results of the research
%    project funded by the National Science Foundation.}

% \tnotetext[2]{The second title footnote which is a longer text matter
%    to fill through the whole text width and overflow into
%    another line in the footnotes area of the first page.}

\author{Laukik Patade}

\ead{laukikpatade22@gmail.com}

\credit{Conceptualization, Programming, Methodology, Writing – original draft}


\author{Rohan Rane}
\ead{ryrane.in@gmail.com}
\credit{Simulation, Programming, Validation,
Designing, Software}
\author{Sandeep Pillai}
\ead{sandeepxpillai@gmail.com}

\credit{Programming, Validation, Writing - original draft.}


% \affiliation[1]
\affiliation{organization={Sardar Patel Institute of Technology},
    addressline={Bhavan's Campus, Andheri West}, 
    city={Mumbai},
    % citysep={}, % Uncomment if no comma needed between city and postcode
    postcode={400058 MH}, 
    country={India}}

\begin{abstract}
This study focuses on optimizing path planning for Unmanned Ground Vehicles (UGVs) in precision agriculture through deep reinforcement learning (DRL) techniques in continuous action spaces. The research begins by reviewing traditional grid-based methods, such as A* and Dijkstra’s algorithms, and highlights their limitations in dynamic agricultural environments, underscoring the need for adaptive learning strategies. Transitioning to DRL approaches, the study explores methods like Deep Q-Networks (DQN), which demonstrate improved adaptability and performance in 2D simulations. Enhancements, including Double Q-Networks and Dueling Networks, are evaluated to further refine decision-making. Building on these insights, the focus shifts to continuous action space models, specifically Deep Deterministic Policy Gradient (DDPG) and Twin Delayed Deep Deterministic Policy Gradient (TD3), which are tested in increasingly complex environments. Experiments conducted in a 3D environment using ROS and Gazebo showcase the effectiveness of continuous DRL algorithms in navigating dynamic agricultural challenges. Notably, the pretrained TD3 agent achieves a 95\% success rate in dynamic environments, demonstrating the robustness of the proposed approach in handling moving obstacles while ensuring no harm to crops or the robot.

\end{abstract}
\begin{highlights}
% \begin{itemize}
    \item This study presents a novel approach to optimizing path planning for Unmanned Ground Vehicles (UGVs) in precision agriculture, employing deep reinforcement learning (DRL) techniques in both discrete and continuous action spaces.
     
    \item Traditional path planning methods like A* and Dijkstra’s algorithms are compared with DRL techniques, highlighting the latter’s superior adaptability in dynamic agricultural environments.
    
    \item The research explores both 2D and 3D environments, with 2D simulations conducted using Pygame and 3D experiments carried out in ROS and Gazebo, where DRL models like Deep Q-Network (DQN), Deep Deterministic Policy Gradient (DDPG), and Twin Delayed Deep Deterministic Policy Gradient (TD3) are evaluated.
    
    \item DRL models, particularly those handling continuous action spaces, demonstrated significant improvements in path efficiency, obstacle avoidance, and real-time adaptability in complex, dynamically changing agricultural fields.
    
    \item The results emphasize the importance of continuous learning algorithms for UGVs in agricultural tasks, allowing more efficient navigation and enhancing performance over traditional algorithms.
    
    \item This study is a part of the larger project "Drones for Smart Agriculture: An Efficient Method for Pesticide Application to Diseased Crops", funded by IEEE, and contributes specifically to optimizing UGVs for pesticide spraying tasks in real-world agricultural scenarios.

\end{highlights}


\maketitle
\section{Statement and Declarations}
\label{declarations}
This research is part of the project titled "Drones for Smart Agriculture: An Efficient Method for Pesticide Application to Diseased Crops" funded by the IEEE Aerospace and Electronic Systems Society (AESS) under the Distributed Sensor Technology and Education Initiative (DSTEI), with a grant of \$25,000. The project comprises three modules:

\begin{enumerate}
    \item \textbf{UAV Module:} A drone monitors crops by capturing images and videos over the agricultural landscape.
    \item \textbf{Image Processing Module:} Captured data is processed at a ground station to detect crop diseases and relay location information to the ground robot.
    \item \textbf{UGV Module:} The ground robot receives the location data, plans a path, and executes targeted pesticide spraying.
\end{enumerate}

\begin{figure*}
    \centering 
    \fbox{\includegraphics[width=0.98\textwidth]{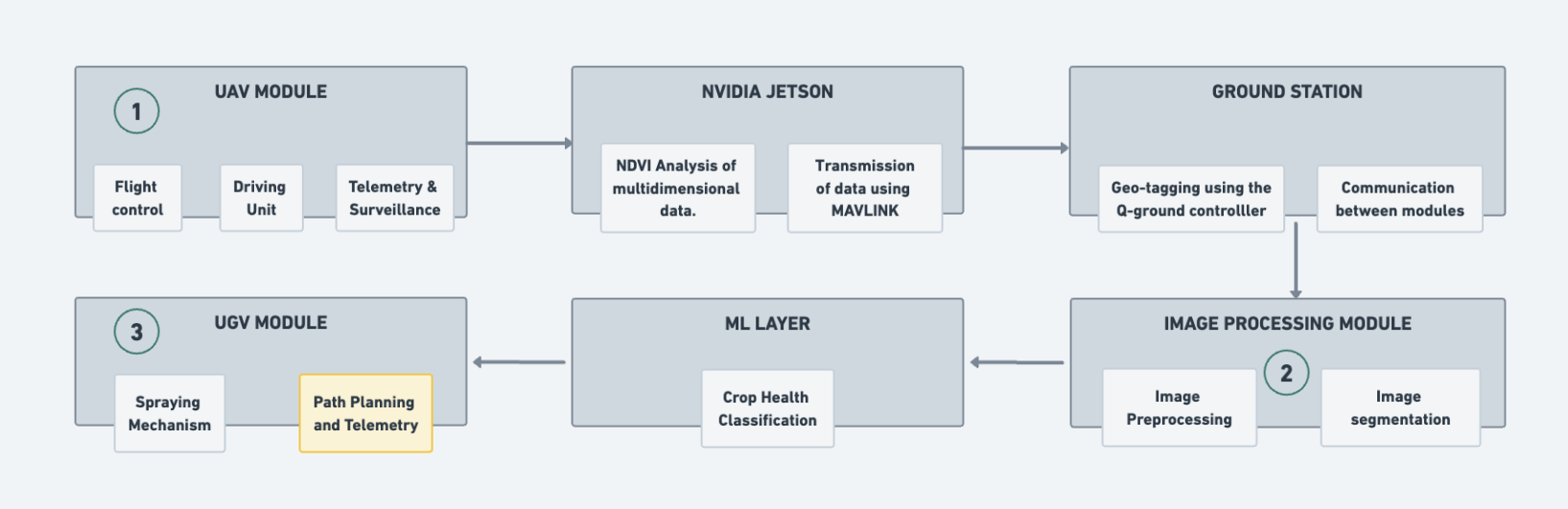}}
    \caption{Drones for Smart Agriculture - Module Diagram} 
    \label{fig:module-diagram}%
\end{figure*}

\noindent
This paper focuses on optimizing the path-planning capabilities of the ground robot, ensuring: 
\begin{enumerate} 
    \item No damage to the robot. 
    \item No harm to surrounding crops. 
    \item Minimized vulnerability to external factors. 
\end{enumerate}
The results contribute directly to the highlighted section in the \cref{fig:module-diagram} module diagram of the overall project.

\section{Introduction}
\label{introduction}

Robotics and Artificial Intelligence (AI) are transforming automation across diverse domains, unlocking unprecedented opportunities for innovation. Among these, the advancements in unmanned systems—particularly Unmanned Aerial Vehicles (UAVs) and Unmanned Ground Vehicles (UGVs)—have been remarkable. Leveraging these advancements, this paper focuses on optimizing the design of UGVs for agricultural applications, with an emphasis on robust path-planning algorithms driven by Deep Reinforcement Learning (DRL). UGVs play a critical role in various fields, including defense, surveillance \citep{GADEKAR202323}, and precision agriculture, where they contribute to operational efficiency and sustainability.

Precision agriculture has significantly benefited from AI-driven technologies that enhance resource allocation and crop management. Studies such as \cite{talaviya2020} underscore how innovations like drones, robotics, and soil sensors optimize irrigation, pest control, and soil preservation, promoting sustainability and improving crop quality. For example, AI-powered systems that dynamically adjust pesticide dosages based on real-time assessments have minimized chemical usage while maximizing pest management efficiency \citep{tewari2020}.

Building on these advancements, our research proposes an adaptive path-planning framework for UGVs operating in agricultural landscapes. This system is designed to complement UAVs, which capture high-resolution crop images for analysis at a central ground station. The processed data informs the UGV's path planning to precisely navigate to diseased crop areas and apply targeted treatments. Such integrated systems have been successfully built before \citep{mammarella2022} and we intend to enhance the performance of UGV subsystems to ensure smooth integration and resource optimization. This would encompass efforts centered exclusively around optimizing UGV navigation and decision-making through DRL.

Traditional path-planning algorithms like A* have proven inadequate in handling the dynamic and uncertain nature of agricultural environments. To address these challenges, this paper investigates the application of DRL as a robust alternative. DRL-based methods allow UGVs to learn and adapt to real-world complexities, such as obstacle-rich terrains and unpredictable conditions. For instance, \cite{ortataŞ2023} demonstrated the utility of machine learning in autonomous navigation, showcasing a GPS-independent UGV equipped with stereo cameras for robust obstacle detection in unstructured settings. Such advancements highlight the potential of ML-based approaches to overcome limitations inherent in conventional path-planning techniques.

Moreover, recent innovations in DRL-based path planning, including prioritized experience replay and expert-driven learning \citep{liu2024}, have further enhanced the efficiency and adaptability of autonomous systems. These methods enable UGVs to optimize navigation strategies dynamically, improving task efficiency in challenging agricultural environments.
This study advances DRL-based path-planning techniques for UGVs to tackle challenges in precision agriculture. By improving autonomous navigation, it contributes to developing intelligent, efficient, and sustainable agricultural systems.

\section{Related Work}
\label{related}

Path planning for mobile robots, including Unmanned Ground Vehicles (UGVs), has been widely researched, exploring various techniques for navigating complex environments. Traditional methods like A* and Rapidly-exploring Random Trees (RRT) are commonly used but face challenges in dynamic settings like precision agriculture. For instance, A* is efficient in static environments \citep{zhi2019}, but struggles with dynamic changes and moving obstacles, which are common in agricultural fields. Frequent path re-computation adds computational overhead, making A* less practical for real-time, dynamic scenarios.

Similarly, RRT-based approaches offer flexibility by exploring the search space randomly \citep{li2021}, but require recalculations when new obstacles arise. While effective in cluttered environments, they struggle to adapt in precision agriculture, where dynamic changes are frequent. Although RRT has been extended for dynamic environments, its need for continuous replanning limits its efficiency in real-time scenarios.

To address these limitations, recent research has shifted towards Deep Reinforcement Learning (DRL), enabling continuous adaptation without repetitive recalculations. The Deep Q-Network (DQN) algorithm \citep{mnih2015} introduced a significant advance by allowing learning from high-dimensional inputs like raw sensory data, crucial in precision agriculture where data is often noisy or incomplete. DQN’s experience replay stabilizes learning by reducing correlations between consecutive observations, enabling UGVs to adapt to real-time changes. \cite{chu2023} applies a double deep Q-network (DDQN) to underwater vehicle navigation, showing effective adaptation in dynamic environments.

Further improving real-time path planning, \cite{yan2020} explored Dueling Double Deep Q-Networks (D3QN) for UAV path planning in dynamic environments. This approach incorporates a situation assessment model to map global threats and a hybrid action selection policy, outperforming DDQN and DQN in cumulative rewards and safe path generation. This demonstrates DRL’s potential for dynamic path planning, not only in UAVs but also in UGVs in similar environments.

DRL has also been applied to multi-robot coordination. Double Q-learning with n-step returns has been used in multi-robot systems in partially known environments, reducing path errors and enhancing efficiency \citep{yang2020}. Studies have also explored DRL for emergency vehicles navigating congested urban roads \citep{yan2021}, highlighting DRL’s adaptability across different applications. This research emphasizes road architecture and vehicle queue dynamics, showcasing DRL’s potential for planning in complex traffic scenarios. A review by \cite{adzhar2020} highlights the advantages of machine learning approaches over classical methods in dynamic environments.

For more complex scenarios, actor-critic methods like Deep Deterministic Policy Gradient (DDPG) and Twin Delayed Deep Deterministic Policy Gradient (TD3) have been applied in continuous action spaces. These techniques are useful in precision agriculture, where environments require continuous actions. For example, \cite{josef2020} applied soft actor-critic methods for local path planning on rough terrain, enabling vehicles to navigate uneven landscapes. Such capabilities are crucial for agricultural UGVs, which must avoid dynamic obstacles. SAC has also been used to enhance training efficiency and handle sparse reward challenges \citep{zhao2023}. In another study, DDPG was applied for rough terrain navigation, allowing UGVs to make real-time decisions based on sensory inputs \citep{guo2020}.

\citep{pengzhan2022} shows that combining SAC with Prioritized Experience Replay (PER) is highly effective for real-time path planning and dynamic obstacle avoidance. This approach outperforms DDPG and A3C in path smoothness, success rates, and computational efficiency. Additionally, \citep{DESHPANDE2024156} proposed integrating DDPG with a Differential Gaming (DG) strategy for exploration in path planning. The DG approach ensures collision-free target reaching by incorporating positive learning episodes into the memory buffer. This strategy improved success rates, reduced collisions, and increased convergence rates, demonstrating the effectiveness of combining exploration strategies like DG with DRL methods for dynamic path planning.

In summary, while classical algorithms like A* and RRT have laid the groundwork for UGV path planning, they struggle in dynamic and unpredictable environments. In contrast, DRL algorithms like DQN, SAC, and actor-critic methods like TD3 offer the adaptability and precision needed for agricultural applications. By enabling continuous learning and adaptation, DRL methods effectively address dynamic obstacle avoidance and real-time decision-making, advancing UGV path planning for precision agriculture.

\section{Methodology}
\label{methodology}
\begin{figure}[h]
    \centering
    \fbox{\includegraphics[width=14cm, height=8cm]{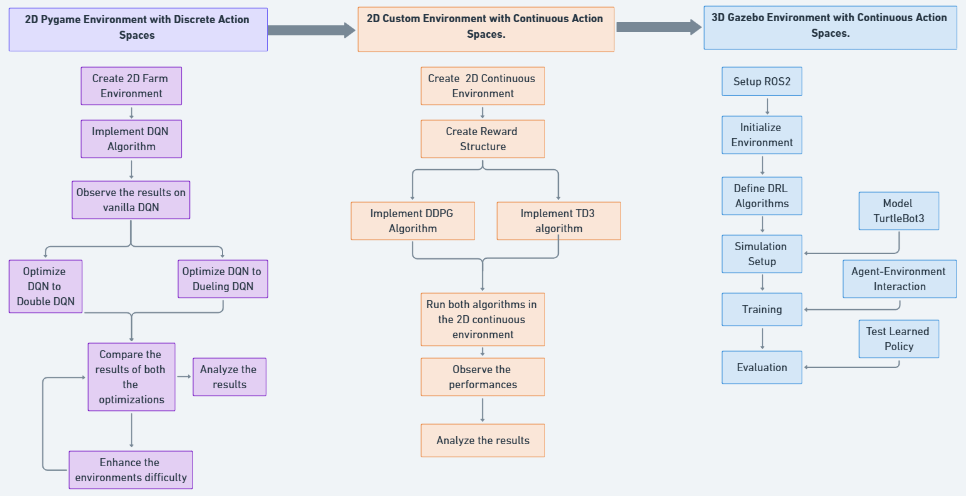}}
    \caption{Methodology Diagram}
    \label{fig:consolidated methodology diagram}
\end{figure}

\noindent
The experimentation is divided into three distinct stages(\cref{fig:consolidated methodology diagram}) to thoroughly analyze the performance of deep reinforcement learning techniques for unmanned ground vehicles (UGVs) in the context of precision agriculture. Each stage systematically explores different action space representations and environmental setups, starting with simpler discrete environments and gradually progressing to more complex and realistic agricultural simulations. This structured approach as outlined in \cref{tab:experiment-stages} allows for a comprehensive evaluation of the algorithms’ effectiveness and adaptability, as they face varying challenges that closely resemble real-world scenarios encountered in agricultural tasks.

\begin{table}[H]
\centering

\caption{Experiment Stages in Path Planning for UGV}
\label{tab:experiment-stages}
\begin{tabular}{|p{3cm}|p{4cm}|p{4cm}|p{4cm}|}
\hline
\textbf{Stage} & \textbf{Environment Setup} & \textbf{Algorithm Implementation} & \textbf{Observations} \\ \hline
\textbf{Stage 1: DQN in Discrete Action Spaces} & 2D grid-based environment with random obstacles simulating agricultural terrain. & DQN for path planning, neural network approximates Q-value function. & Track episode length, rewards, and convergence speed. Compare DQN, Double DQN, and Dueling DQN. \\ \hline
\textbf{Stage 2: Continuous Action Spaces} & Extend 2D grid to continuous movement, allow varying speeds and angles. & DDPG and TD3 (Actor-Critic methods) for continuous action space. & Evaluate rewards, episode length, and training stability.  \\ \hline
\textbf{Stage 3: Realistic Agricultural Simulation} & 3D agricultural environment in Gazebo with crops, rocks, and uneven terrain. & Integrate optimized model in Gazebo for path planning. & Track success rate, rewards, and training loss. Evaluate model's effectiveness in a real scenario. \\ \hline
\end{tabular}

\end{table}

\section{Experimentation - Discrete Action Spaces}

\subsection{Deep Q-Network}
\vspace{1em}
\subsubsection{Environment}
\begin{figure}[h]
    \centering
    \fbox{\includegraphics[width=6cm,height=6
    cm]{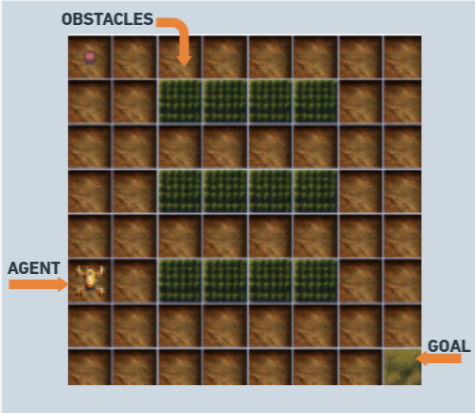}}
    \caption{2D Farm Environment in Pygame}
    \label{fig:2D Farm Environment}
\end{figure}

\noindent
The environment is built on top of the existing FrozenLake - v1 environment \citep{frozenlake} of the gymnasium package. A replica of a 2D grid-based agricultural landscape is constructed as shown in \cref{fig:2D Farm Environment}.

\vspace{1em}
\noindent
There are 4 entities within the environment, namely:
\begin{enumerate}
    \item The agent
    \item Obstacles
    \item Goal
    \item Free space
\end{enumerate}

\begin{center}
The agent can be in 64 states and can take 4 actions (UP, DOWN, RIGHT, LEFT).
\end{center}

% {2}

\subsubsection{Reward Structure}
\noindent
The reward structure utilized in this experiment mirrors that of the OpenAI Gym environment, which is characterized as a sparse reward structure. This type of reward shaping assigns rewards infrequently, primarily focusing on significant milestones rather than providing continuous feedback for every action taken. The rewards are assigned as follows:

\begin{itemize} 
    \item Reaching the goal crop: +1 
    \item Colliding with an obstacle: 0 (terminates the episode) 
    \item Reaching an open field: 0 
\end{itemize}

\noindent
In this context, the maximum achievable reward for each episode is 1. It should be noted that the robot is not penalized for its actions; instead, episodes are terminated when the robot is struck by an obstacle. The designation of "sparse" arises from the limited number of reward signals provided, which encourages the agent to explore the environment more extensively before receiving feedback. 

\subsubsection{Deep Q-Network (DQN) Architecture}
\noindent
The \nameref{Deep Q-Network (DQN) algorithm} \citep{volodymyrmnih2015} extends Q-learning by using a neural network to approximate Q-values, aiming to maximize cumulative reward through balanced exploration and exploitation.

\paragraph{Policy Network (Q-network)}:
The policy network, \( Q(s, a|\theta) \), estimates Q-values for actions in a given state \( s \) and is updated via gradient descent. Actions are selected using epsilon-greedy: with probability \( \epsilon \), a random action is chosen (exploration); otherwise, the action with the highest Q-value is selected (exploitation).

\paragraph{Target Network}:
The target network \( Q'(s, a|\theta^-) \) is a periodically updated copy of the policy network with weights \( \theta^- \). This target network stabilizes training by providing consistent Q-values for target calculations. Instead of updating with every policy change, the target network is updated less frequently, helping to smooth out learning and prevent oscillations.

\paragraph{Replay Memory}:
In replay memory, transitions \( (s_t, a_t, r_t, s_{t+1}, \text{terminated}) \) are stored in a buffer. Mini-batch sampling from this buffer breaks the sequential correlations between transitions. By enabling the agent to learn from a diverse set of past experiences, replay memory enhances generalization and robustness.

\paragraph{Epsilon-Greedy Policy}:
The epsilon-greedy policy manages the trade-off between exploration and exploitation. With probability \( \epsilon \), the agent selects a random action, promoting exploration of new strategies. Over time, \( \epsilon \) decays, encouraging the agent to increasingly favor the optimal action and leverage its learned knowledge to maximize performance. This controlled decay in \( \epsilon \) allows for systematic exploration in the initial stages and refined action selection later on.

\paragraph{Bellman Equation and Loss Function}:
The target Q-value \( y_i \) is computed as:
\[
y_i = 
\begin{cases} 
r_i & \text{if terminal,} \\
r_i + \gamma \max_a Q'(s_{i+1}, a|\theta^-) & \text{otherwise.}
\end{cases}
\]
The Bellman equation describes this recursive target relationship, capturing how expected rewards propagate over time. The loss \( L(\theta) \), calculated as the mean squared error between \( y_i \) and \( Q(s_i, a_i|\theta) \), minimizes the difference between predicted and target values, allowing the network to more accurately estimate the optimal Q-values.

\paragraph{Experience Replay}:
Experience replay further stabilizes training through random mini-batch sampling from stored transitions in replay memory. By allowing the agent to revisit past experiences, it mitigates the impact of highly correlated and sequential data. This process enables better representation of the state-action space by leveraging a broad array of past interactions.

\paragraph{Target Network Update}:
The target network periodically receives updated weights from the policy network, \( \theta^- \leftarrow \theta \), allowing it to reflect newer learning without the instability caused by frequent updates. This delayed update provides a consistent target for multiple steps, enhancing stability by smoothing the learning trajectory.

\vspace{1em} 
\noindent
This approach encourages the agent to explore the environment through random actions in the initial episodes, balancing exploration and learning. As the episodes progress and \( \epsilon \) decays, the agent focuses on maximizing cumulative rewards by leveraging the knowledge gathered from earlier exploration.

\begin{algorithm}[h]
\caption{Deep Q-Network (DQN) Algorithm}
\label{Deep Q-Network (DQN) algorithm}
\begin{algorithmic}[1]
\State Initialize replay memory $R$ with capacity $N$
\State Initialize policy network $Q(s, a|\theta)$ with random weights
\State Initialize target network $Q'(s, a|\theta^-)$ with weights copied from the policy network
\State Set learning rate $\alpha$, discount factor $\gamma$, and exploration rate $\epsilon$
\For{each episode}
    \State Reset the environment and get the initial state $s_0$
    \For{each step $t$ in the episode}
        \State With probability $\epsilon$, select a random action $a_t$
        \State Otherwise, select $a_t = \arg\max_a Q(s_t, a|\theta)$
        \State Execute action $a_t$ and observe reward $r_t$ and new state $s_{t+1}$
        \State Store transition $(s_t, a_t, r_t, s_{t+1}, \text{terminated})$ in replay memory $R$
        
        \If{replay memory $R$ has enough samples}
            \State Sample a mini-batch of transitions $(s_i, a_i, r_i, s_{i+1}, \text{terminated}_i)$ from $R$
            \For{each sampled transition}
                \If{$\text{terminated}_i$ is True}
                    \State Set target $y_i = r_i$
                \Else
                    \State Set target $y_i = r_i + \gamma \max_a Q'(s_{i+1}, a|\theta^-)$
                \EndIf
                \State Compute the loss: 
                \begin{equation}
                L = \frac{1}{N} \sum_i \left( y_i - Q(s_i, a_i|\theta) \right)^2
                \end{equation}
            \EndFor
            \State Perform a gradient descent step to minimize the loss
        \EndIf
        \State Update state $s_t \leftarrow s_{t+1}$
        \If{step $t$ is a target update step}
            \State Update target network weights $\theta^- \leftarrow \theta$
        \EndIf
    \EndFor
    \State Decay exploration rate $\epsilon$
\EndFor
\end{algorithmic}
\end{algorithm}

\subsubsection{Neural Network Architecture}
\noindent
The Deep Q-Network (DQN) consists of the following key elements:

\begin{itemize}
    \item \textbf{Input dimensions}: Match the environment's state space, representing the robot’s current situation in terms of position, orientation, or sensor data. For an 8x8 grid environment, the input is a binary representation of the 64 possible states. 
    
    \item \textbf{Output dimensions}: Correspond to the action space, with each output providing a Q-value for one possible action. In the 8x8 grid, the output includes Q-values for all four state-action pairs associated with the current state.(UP, LEFT, RIGHT, BOTTOM)
    
    \item \textbf{Input to the network}: The current state, represented as a feature vector (binary for the 8x8 grid), enables the DQN to calculate Q-values based on the robot's environment.  
    
    \item \textbf{Output of the network}: Provides Q-values for all actions, with the highest Q-value corresponding to the action selected during decision-making.  
    
    \item \textbf{Activation function}: ReLU (Rectified Linear Unit) is used in hidden layers to introduce non-linearity, enhancing the network’s capacity to learn complex patterns.  
\end{itemize}

% {3}
\begin{figure}
    \centering
    \fbox{\includegraphics[width=10cm,height=7.5cm]{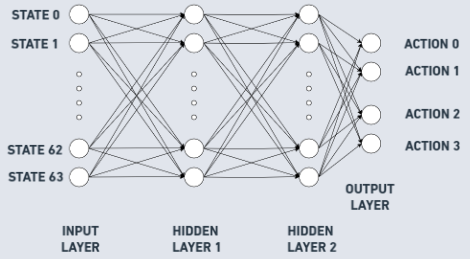}}
    \caption{DQN Neural Network Architecture}
    \label{fig:DQN NN}
\end{figure}

\noindent
The network as shown in \cref{fig:DQN NN} consists of an input layer, two hidden layers with ReLU activations, and an output layer. Both the policy and target networks share this architecture, and the replay buffer stores the agent's past experiences, which are later sampled for training. The hyperparameters for both, the policy and the target network are tabulated in \cref{tab:dqn_hyperparameters}.

\begin{table}[h]
    \captionsetup{justification=centering, font=small, labelsep=period, skip=0.5em} % Adjust caption properties
    \caption{Hyperparameters - Deep Q Network (DQN)}
    \label{tab:dqn_hyperparameters}
    \centering
    \small % Reduces font size to fit better in the column
    \begin{tabular}{|p{3.0cm}|p{1.0cm}|p{3.2cm}|} \hline 
    
        \textbf{Hyperparameter} & \textbf{Value} & \textbf{Description} \\ \hline 
        Learning Rate & 0.001 & Parameter update rate \\ \hline 
        Discount Factor ($\gamma$) & 0.99 & Future reward discount factor \\ \hline 
        Epsilon Decay Rate & 0.995 & Exploration decay rate \\ \hline 
        Minimum Epsilon ($\epsilon$) & 0.01 & Minimum exploration rate \\ \hline 
        Replay Buffer Size & 100,000 & Max capacity for experiences \\ \hline 
        Batch Size & 64 & Samples per update \\ \hline 
        Target Network Update Freq. & 1000 & Steps between target updates \\ \hline 
        
    \end{tabular}
\end{table}

\subsubsection{Performance}

\begin{figure}[h]
    \centering
    % First subfigure
    \begin{subfigure}[t]{0.48\textwidth} % Adjust the width as needed
        \centering
        \fbox{\includegraphics[width=6.5cm,height=6.5cm]{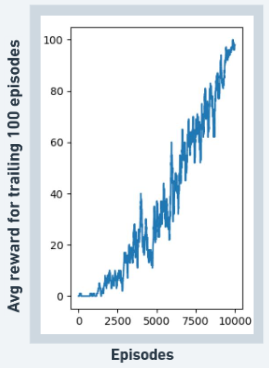}}
        \caption{Trailing 100 Rewards vs Episodes (DQN)}
        \label{fig:Rewards DQN}
    \end{subfigure}
    \hfill
    % Second subfigure
    \begin{subfigure}[t]{0.48\textwidth} % Adjust the width as needed
        \centering
        \fbox{\includegraphics[width=6.5cm,height=6.5cm]{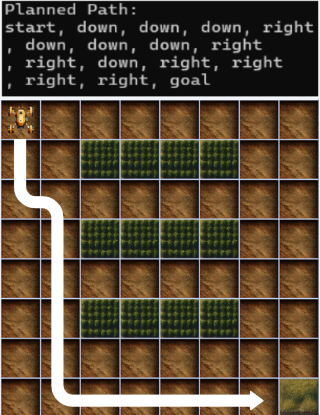}}
        \caption{Planned Path}
        \label{fig:Planned Path 8x8}
    \end{subfigure}
    \caption{Performance of DQN in 8X8 Environment}
    \label{fig:combined}
\end{figure}

\noindent
The neural network is trained to optimize the approximation of target Q-values for 10,000 steps, with evaluation based on the trailing rewards over the last 100 episodes. The model is considered to have converged when it consistently reaches the goal in all 100 consecutive episodes. The illustration \cref{fig:Planned Path 8x8} shows that the agent reaches the goal in the minimum number of steps, selecting one of the optimal paths. However, similar results can be achieved by simple grid-based algorithms like A* and Dijkstra's algorithm. But these algorithms lack the capacity to learn or adapt to environmental dynamics, which DQN does. This can be demonstrated by introducing noise into the environment. The FrozenLake-v1 environment includes an external disturbance parameter called \texttt{is\_slippery}, which when enabled, causes the agent to deviate from the most optimal action.

% {4}

\noindent
A significant difference is observed in \cref{fig:Noisy Rewards} during training in the noisy environment, where the agent took 40,000 episodes to achieve 80\% convergence. This implies that the agent reaches the goal in 8 out of 10 attempts on average. Testing in the noisy environment:

\begin{figure}[h]
    \centering
    % First subfigure
    \begin{subfigure}[t]{0.48\textwidth}
        % {6A} % Customize figure numbering
        \centering
        \fbox{\includegraphics[width=6.5cm,height=6.5cm]{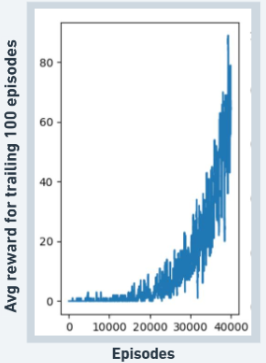}}
        \caption{Trailing 100 Rewards vs Episodes (Noisy Env)}
        \label{fig:Noisy Rewards}
    \end{subfigure}
    \hfill
    % Second subfigure
    \begin{subfigure}[t]{0.48\textwidth}
        % {6B} % Customize figure numbering
        \centering
        \fbox{\includegraphics[width=6.5cm,height=6.5cm]{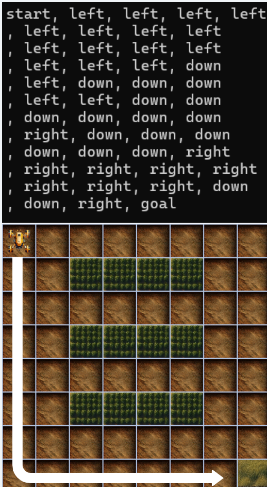}}
        \caption{Planned Path in Noisy Environment}
        \label{fig:Planned Path Noisy}
    \end{subfigure}
    \caption{Performance of DQN in 8x8 Noisy Environment}
    \label{fig:combined Noisy}
\end{figure}

\noindent
\cref{fig:Planned Path Noisy} demonstrates that the agent requires significantly more steps due to the \texttt{is\_slippery} factor, which continuously pushes it off course. By learning the dynamics of the noisy environment, the agent plans its path along the boundary to avoid obstacles---a clear sign of adaptive intelligence. This is especially valuable in complex environments like agricultural fields, where many disturbances cannot be quantified. Unlike simple algorithms such as A* or RRT, which rely solely on predefined paths, DQN’s ability to learn and adapt from experience makes it more effective in such dynamic landscapes.

\subsection{Towards improving DQN}
\noindent
While the basic Deep Q-Network (DQN) algorithm provides a solid foundation for path planning in discrete environments, it is essential to implement further optimizations to enhance its performance in precision agriculture tasks. The goal of these optimizations is to improve the robustness, speed, and reliability of DQN-based Unmanned Ground Vehicle (UGV) navigation.

\subsubsection{Double DQN}
\noindent
One significant improvement is the introduction of Double DQN \citep{hasselt2016}, which addresses the issue of maximization bias inherent in the original DQN algorithm. In standard DQN, the action selection and action evaluation both happen with the target network 
$Q'(s_{i+1}, a'; \theta^-)$. This can lead to an overestimation of Q-values. It selects the action $a'$ that maximizes $Q'$, and then evaluates this action using the same network, which can cause over-optimistic estimates of the action's value. Double DQN mitigates this bias by simply using the policy network for selecting the best action and the target network to evaluate its Q-value, which results in a more stable and accurate update process. Action selection is done using the current policy network $Q(s_{i+1}, a'; \theta)$, 
but the evaluation of that action's value is done using the target network $Q'(s_{i+1}, a'; \theta^-)$. This decoupling reduces the risk of overestimating Q-values because the action selection and value evaluation are not based on the same function. In standard DQN, the target Q-value update rule is:
\[
y_i^{DQN} = r_i + \gamma \max_{a'} Q'(s_{i+1}, a'; \theta^-)
\]

In Double DQN, the update rule is modified to decouple action selection and action evaluation:
\[
y_i^{Double DQN} = r_i + \gamma Q'(s_{i+1}, \arg\max_{a'} Q(s_{i+1}, a'; \theta); \theta^-)
\]

Thus, instead of just selecting the state-action pair with the maximum Q value, we use the target network to evaluate the Q value and determine the optimal action. This leads to better Q-value estimation.

\subsubsection{Dueling Networks}
\begin{figure}[h]
    \centering
    \fbox{\includegraphics[width=8.5cm,height=5.7cm]{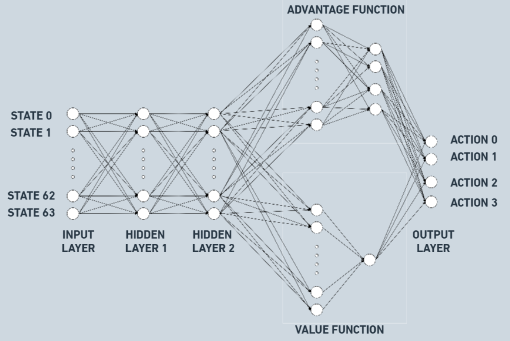}}
    \caption{Dueling Networks Architecture}
    \label{fig:D3QN NN}
\end{figure}

\noindent
Dueling DQN \citep{wang2016} introduces a structural change (shown in \cref{fig:D3QN NN}) where the estimation of the Q-value is split into two components: the value function and the advantage function. This separation helps the network learn more efficiently by focusing on two distinct aspects:

\begin{itemize}
    \item \textbf{State Value} \( V(s) \): This represents how valuable it is to be in a particular state, regardless of the action taken.
    
    \item \textbf{Advantage} \( A(s, a) \): This represents how advantageous it is to take a specific action in a given state, relative to other actions.   
\end{itemize}

\noindent
By splitting the network into these two streams, Dueling DQN allows the network to better evaluate which states are inherently valuable and which actions are preferable. This separation is particularly useful in environments where many actions may have similar outcomes in a given state, but certain states are more critical than others.

% \vspace{0.5em} 
The Q-value is then computed by combining the value and advantage streams:
\[
Q(s, a) = V(s) + \left( A(s, a) - \frac{1}{|\mathcal{A}|} \sum_{a'} A(s, a') \right)
\]

\noindent
This formulation helps reduce the impact of noisy or redundant action-value estimates by normalizing the advantage across actions. As a result, D3QN can more effectively prioritize learning the value of different states, even in situations where the action choices may not differ significantly. This modification enhances the agent's ability to differentiate between important states and actions, leading to improved performance, especially in complex environments where actions might not significantly vary in value across states.

\subsection{Observations}

\subsubsection{8x8 Environment}
\noindent
Following the successful implementation of Double DQN and Dueling Double DQN, the environment was run on these new variants to observe their performance outcomes.

\begin{figure}[h]
    \centering
    % First subfigure
    \begin{subfigure}[t]{0.45\textwidth}
        \centering
        \fbox{\includegraphics[width=6cm,height=6cm]{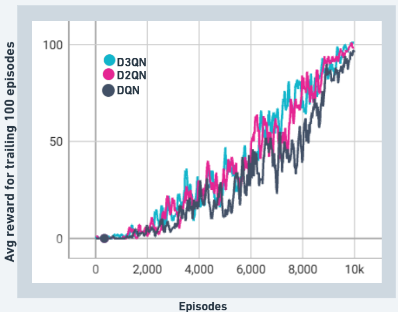}}
        \caption{DQN vs D2QN vs D3QN - Trailing Rewards}
        \label{fig:dqn-d2qn-d3qn}
    \end{subfigure}
    \hfill
    % Second subfigure
    \begin{subfigure}[t]{0.45\textwidth}
        \centering
        \fbox{\includegraphics[width=6cm,height=6cm]{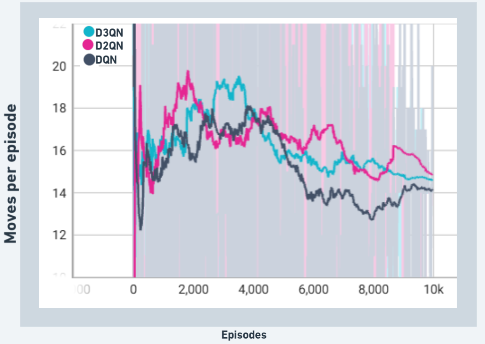}}
        \caption{DQN vs D2QN vs D3QN: Moves per Episode}
        \label{fig:dqn-d2qn-d3qn-mov-per-episode}
    \end{subfigure}
    \caption{Performance of DQN, D2QN \& D3QN in 8x8 Environment}
    \label{fig:comparison-dqn-d2qn-d3qn}
\end{figure}

% \vspace{1em} 
The analysis of \cref{fig:dqn-d2qn-d3qn} and \cref{fig:dqn-d2qn-d3qn-mov-per-episode} \textit{(*smoothened by a factor of 0.99)} indicates that in a simple environment like the 8x8 agricultural map, all variants of the DQN algorithm demonstrate similar performance metrics. Notably, Dueling DQN exhibits slightly higher efficiency concerning the number of moves per episode, although the differences are not substantial. The straightforward nature of the map facilitates efficient learning for the DQN algorithm, enabling all optimized variants to reach comparable performance thresholds. Consequently, modifications to the environment are warranted, transitioning to a more complex 10x10 map to further evaluate the algorithms' capabilities.

\subsubsection{10x10 Environment}
\begin{figure}[h]
    \centering
    % First subfigure
    \begin{subfigure}[t]{0.45\textwidth}
        \centering
        \fbox{\includegraphics[width=5.5cm,height=5.5cm]{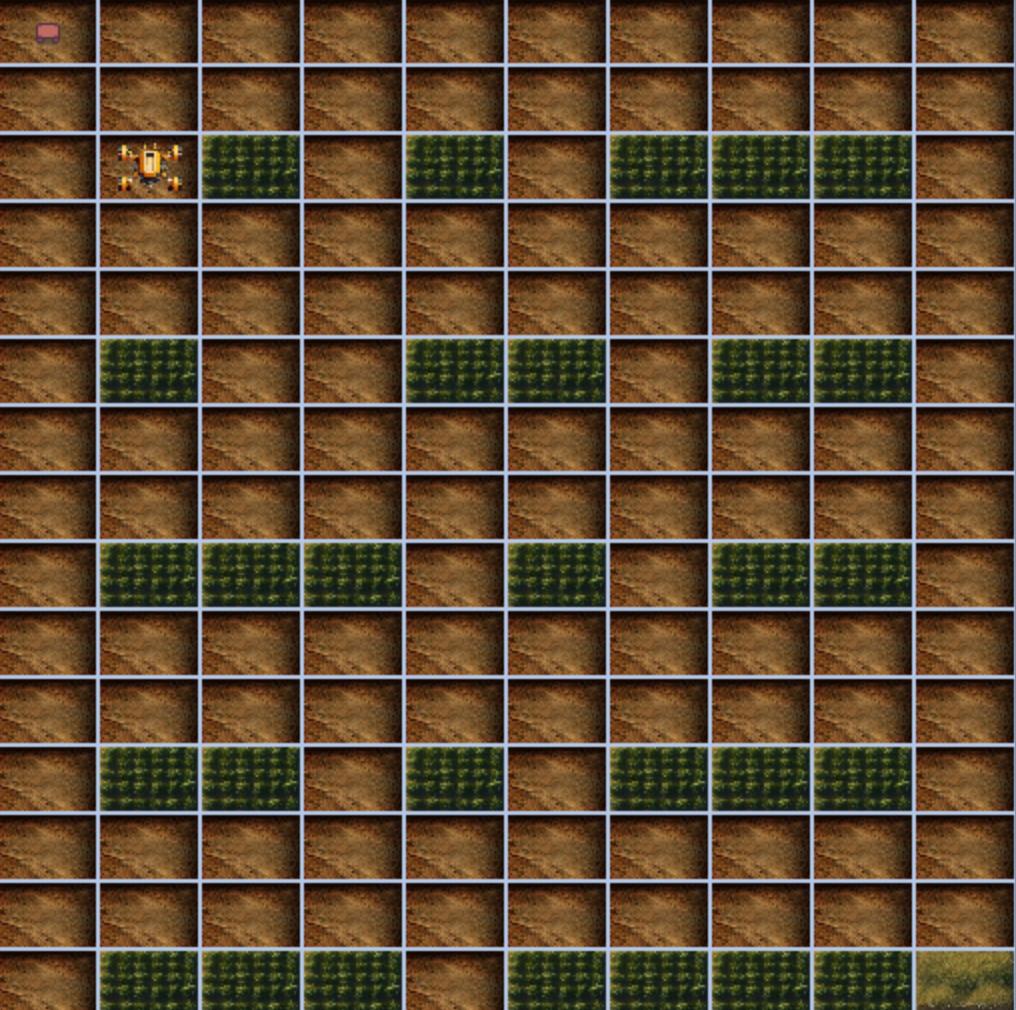}}
        \caption{10x10 Map}
        \label{fig:10x10-map}
    \end{subfigure}
    \hfill
    % Second subfigure
    \begin{subfigure}[t]{0.45\textwidth}
        \centering
        \fbox{\includegraphics[width=5.5cm,height=5.5cm]{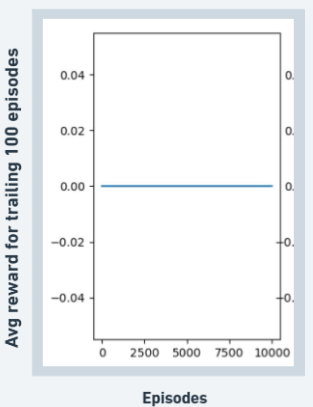}}
        \caption{DQN Performance in 10x10 Map}
        \label{fig:dqn-10x10}
    \end{subfigure}
    \caption{Visualization of the 10x10 map and corresponding DQN performance. (A) Map layout and (B) DQN results.}
    \label{fig:10x10-comparison}
\end{figure}

\noindent
To thoroughly assess the optimization performance of the various algorithmic variants in a more complex environment compared to the simpler 8x8 grid, a new 10x10 grid (\cref{fig:10x10-map}) was meticulously designed. This grid features a significantly increased number of obstacles, making the environment computationally more demanding and challenging for the agent. Despite this added complexity, all parameters within the map remain consistent with those used in the previous environment, with the sole alteration being the layout itself. This new configuration introduces more non-uniform obstacles, thereby enhancing the intricacy of the navigation task that the agent must successfully overcome.

\vspace{0.5em}
\noindent
The following three critical parameters were fixed for this experiment, applicable to DQN, Double DQN, and Dueling DQN:

\begin{itemize}
    \item Environment: 10x10 agricultural map
    \item Episodes: 10,000
    \item Learning Rate: 1e-3
\end{itemize}

\noindent
\textbf{DQN}: Upon executing the DQN algorithm in the 10x10 map, it was observed that, despite the fixed episode count, the agent struggled to learn the map and achieve any rewards.

The agent continuously opted for random actions throughout the map, failing to reach its goal and, consequently, obtaining no rewards (as seen in \cref{fig:dqn-10x10}). Thus, it is evident that DQN underperforms in environments with increased complexity.

\vspace{1em}
\noindent
\textbf{Double DQN}: Using the double DQN algorithm to navigate the environment, a significant improvement over the traditional DQN approach was clearly observed. The agent demonstrated a marked enhancement in its ability to select random actions strategically, allowing it to effectively reach its designated goal while simultaneously collecting rewards more efficiently. 

\begin{figure}[h]
    \centering
    % First subfigure
    \begin{subfigure}[t]{0.45\textwidth}

        \centering
        \fbox{\includegraphics[width=5.5cm, height=5.5cm]{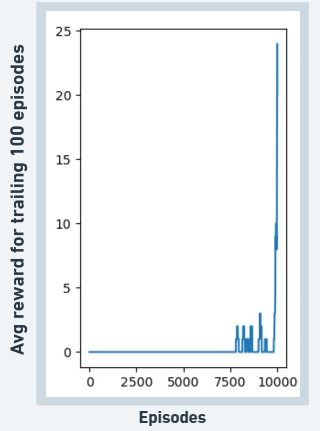}}
        \caption{Double DQN Performance in 10x10 Map}
        \label{fig:d2qn-10x10}
    \end{subfigure}
    \hfill
    % Second subfigure
    \begin{subfigure}[t]{0.45\textwidth}

        \centering
        \fbox{\includegraphics[width=5.5cm, height=5.5cm]{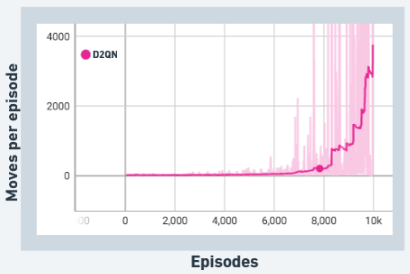}}
        \caption{Moves per Episode for Double DQN in 10x10 Map}
        \label{fig:moves-per-episode-d2qn}
    \end{subfigure}
    \caption{Double DQN Performance in the 10x10 map: (A) Performance and (B) Moves per Episode.}
    \label{fig:d2qn-comparison}
\end{figure}

It can be seen from \cref{fig:d2qn-10x10} that the double DQN algorithm is capable of accumulating intermittent rewards through numerous random actions; however, it has not fully learned the environment. A careful examination of \cref{fig:moves-per-episode-d2qn} \textit{ (* smoothed by a factor of 0.99)} reveals that, although the algorithm is nearing a learning threshold, the frequency of random actions taken remains significantly high. Hence, this approach is inefficient for path planning unless the episode count is increased, providing the algorithm with additional time to learn.

\begin{figure}[h]
    \centering
    % First subfigure
    \begin{subfigure}[t]{0.32\textwidth}

        \centering
        \fbox{\includegraphics[width=5cm, height=5cm]{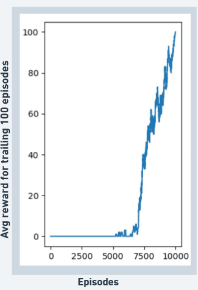}}
        \caption{Dueling DQN Performance in 10x10 Map}
        \label{fig:d3qn-10x10}
    \end{subfigure}
    \hfill
    % Second subfigure
    \begin{subfigure}[t]{0.32\textwidth}

        \centering
        \fbox{\includegraphics[width=5cm, height=5cm]{Moves_per_epi_10X10_D2QN.png}}
        \caption{Moves per Episode for Dueling DQN in 10x10 Map}
        \label{fig:moves-per-episode-d3qn}
    \end{subfigure}
    \begin{subfigure}[t]{0.32\textwidth}

        \centering
    \fbox{\includegraphics[width=5cm, height=5cm]{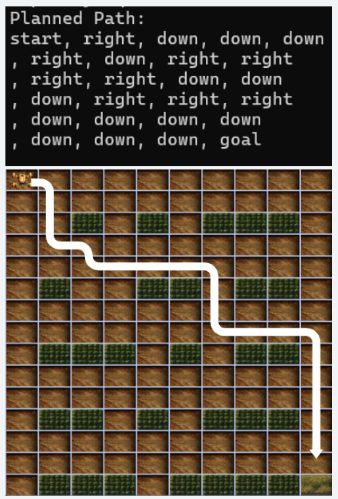}}
    \caption{Path Planned by D3QN for 10X10 Grid}
    \label{fig:Path Planned by D3QN for 10X10 Grid}
    \end{subfigure}
    \caption{Dueling DQN Performance in the 10x10 map: (A) Performance and (B) Moves per Episode.}
    \label{fig:d3qn-comparison}
\end{figure}

\noindent
\textbf{Dueling DQN}: Observing Dueling DQN's performance demonstrates that continued optimization and the rectification of shortcomings observed in previous variants yield greater efficiency and reliability in path learning within a reduced time frame. The agent starts learning around the 5000th episode and is able to fully converge within 10000 episodes. According to \cref{fig:moves-per-episode-d3qn} \textit{(*smoothened by a factor of 0.99)}, Dueling DQN also requires significantly lower amount of random moves to start learning the environment.

To further illustrate the impact of this efficiency, \cref{fig:Path Planned by D3QN for 10X10 Grid} visualizes a path planned by the Dueling DQN in a 10x10 grid. The planned path demonstrates a well-optimized route with minimal detours, indicating a high level of learning stability. This structured path is the result of Dueling DQN's capacity to accurately distinguish between states of higher and lower values, reducing the computational load associated with unnecessary exploration.

\subsection{Results \& Analysis}
\noindent
The experimental results tabulated in \cref{tab:performance_comparison_algorithms} indicate that in a relatively simple environment, such as the 8x8 grid, DQN, D2QN, and D3QN exhibit comparable performance in terms of convergence and average trailing reward. However, both D2QN and D3QN demonstrate noticeable improvements in training time, with D3QN achieving the fastest results.

\noindent
In contrast, within a more complex environment like the 10x10 grid with dense obstacle placement, the optimizations present in D2QN and D3QN result in significant advantages over the baseline DQN. In this scenario, D3QN successfully learns the environment within 10,000 steps, attaining an average trailing reward of 100, indicating complete mastery of the task. D2QN performs moderately, with a final trailing reward of 25, reflecting that the agent reached its goal in only 25 out of 100 episodes, corresponding to a 25\% success rate. In comparison, DQN fails to learn the environment within the same timeframe.

The maximum number of moves taken per episode provides insights into the exploration behavior of each algorithm. D3QN consistently requires fewer moves than both DQN and D2QN, suggesting that the separation of advantage and value networks in D3QN facilitates more efficient exploration, reducing unnecessary actions while still achieving optimal outcomes.

\begin{table}[h]
    \caption{Comparison of DQN, D2QN, and D3QN across two scenarios (8x8 and 10x10)}
    \label{tab:performance_comparison_algorithms}
    \centering
    \renewcommand{\arraystretch}{1.4} % Adjust row height for better spacing
    \setlength{\tabcolsep}{6pt} % Adjust cell padding for better spacing
    \begin{tabular}{|c|c|c|c|c|c|c|}
        \toprule
        \multirow{2}{*}{\textbf{Performance Metric}} & \multicolumn{3}{c|}{\textbf{8x8}} & \multicolumn{3}{c|}{\textbf{10x10}} \\
        \cmidrule(lr){2-4} \cmidrule(lr){5-7}
        & \textbf{DQN} & \textbf{D2QN} & \textbf{D3QN} & \textbf{DQN} & \textbf{D2QN} & \textbf{D3QN} \\
        \midrule
        Convergence Starts                  & 3000        & 3000        & 3000        & -       & 7500        & 5000        \\
        Avg. Trailing Reward after 10k Episodes & 100         & 100         & 100         & 0       & 25          & 100         \\
        Maximum Exploration (Unsmoothened) & 84          & 70          & 64          & 130     & 50000       & 1750        \\
        Training Time (min)                & 5:32        & 4:21        & 3:16        & -       & 16:20       & 8:01        \\
        \bottomrule
    \end{tabular}
\end{table}

\section{Experimentation - Continuous Action Spaces} 
\noindent
While the discrete action space model offers a simplified solution to the path planning problem, it is limited in its ability to fully capture the complexities of real-world agricultural environments. In practice, Unmanned Ground Vehicles (UGVs) operating in the field must have greater degrees of freedom to adapt to dynamic and unstructured terrains. Agricultural environments, with their irregular landscapes, obstacles like trees and crops, and the need for precise navigation, demand more flexible control over movement—something discrete actions cannot adequately provide.

\subsection{\textbf{Environment}}
\begin{figure}[h]
    \centering
    \fbox{\includegraphics[width=12cm, height=5cm]{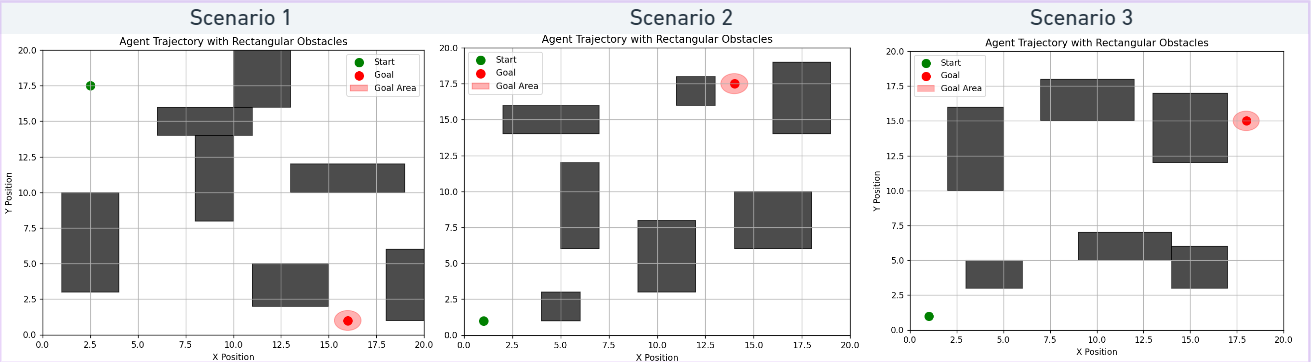}}
    \caption{2D Continuous Environment}
    \label{fig:2-D Continuous Environment}
\end{figure}

\noindent
In this experiment we have leveraged matplotlib to generate 3 scenarios (\cref{fig:2-D Continuous Environment}) in which the agent will be trained.

The environment has the following entities:
\begin{enumerate}
    \item Black spaces which indicate obstacles.
    \item Green filled circle which indicates the starting point of the agent.
    \item Red filled circle which indicates the goal of the agent.
    \item Red translucent circle which indicates the goal area upon reaching which terminates the current episode.
\end{enumerate}

% \vspace{0.5em} 
\noindent
\textbf{Action Space}: The agent can freely move in the environment with a continuous action space of [-1,1].

% \vspace{0.5em} 
\noindent
\textbf{Observation Space}: The agent can be present at any co-ordinates within the region and thus has an observation space of [0,20].

\subsection{\textbf{Reward Shaping}}
\noindent
As demonstrated by \cite{grzes2017}, potential-based reward shaping provides a framework for enhancing agent learning efficiency by shaping the reward structure without changing the underlying optimal policy. 

% \vspace{0.5em} 
In this implementation, the reward structure(\cref{fig:Rewards Structure}) for the continuous environment is as follows:

\begin{itemize}
    \item The agent incurs a penalty of -2 for colliding with or approaching obstacles.
    \item A penalty is applied for navigating too close to the boundary.
    \item Significant penalties are given for revisiting the same states to promote exploration.
    \item A small penalty is incurred for each step taken, encouraging quicker convergence.
    \item The agent receives a dynamic reward based on the Potential-Based Reward Shaping Function, defined as:
    \[
    R_t = -\frac{\text{current distance from the goal}}{\text{distance from the start to the goal}},
    \]
    thereby incentivizing the agent to minimize its distance to the goal with each action.
\end{itemize}

This approach is crucial for real-world applications, where unmanned ground vehicles (UGVs) must continuously adjust their actions to navigate dynamic environments effectively. By employing actor-critic algorithms like DDPG and TD3, specifically designed for continuous action spaces, we can enhance the UGV's performance in precision agriculture scenarios.

\subsection{Actor-Critic Methods}
\begin{figure}[h]
    \centering
    % Subfigure for Reward Structure
     \begin{subfigure}[b]{0.45\textwidth}
        \centering
        \fbox{\includegraphics[width=7cm, height=7cm]{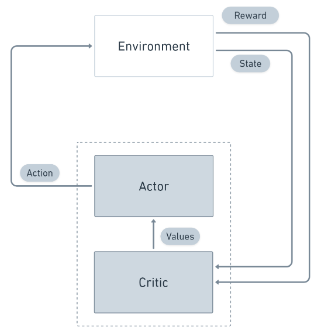}}
        \caption{Actor-Critic Architecture}
        \label{fig:Actor-Critic Architecture}
    \end{subfigure}
    \hfill
    \begin{subfigure}[b]{0.45\textwidth}
        \centering
        \fbox{\includegraphics[width=7cm, height=7cm]{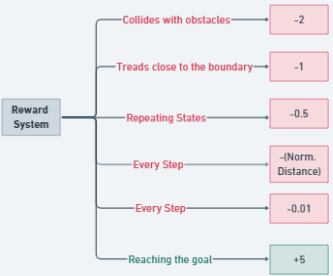}}
        \caption{Reward Structure}
        \label{fig:Rewards Structure}
    \end{subfigure}
    
    % Subfigure for Actor-Critic Architecture
   
    \caption{Visualization of Reward Structure and Actor-Critic Architecture}
    \label{fig:Reward and Actor-Critic}
\end{figure}

\noindent
Actor-Critic methods \citep{grondman2012} are reinforcement learning techniques that combine methodologies of policy gradient techniques and value based methods to optimize the agent’s action in the environment. The Actor-Critic architecture as presented in \cref{fig:Actor-Critic Architecture} is a foundational framework in reinforcement learning that combines two components: an actor and a critic. \textbf{Actor-Critic algorithms} combine two key components in reinforcement learning: the \textbf{actor} and the \textbf{critic}.

\begin{itemize}
    \item \textbf{Actor}: Responsible for selecting actions. It learns a policy \( \pi(s) \), which maps states \( s \) to actions \( a \), and aims to maximize the expected cumulative reward.
  
    \item \textbf{Critic}: Evaluates the actions taken by the actor by estimating the value function \( V(s) \) or the action-value function \( Q(s, a) \). This feedback helps the actor improve its policy by learning which actions yield better outcomes.
\end{itemize}

\vspace{-1em}
\textbf{Key elements}:
\begin{enumerate}
    \item \textbf{Policy-based (Actor)}: The actor updates the policy directly using gradients (policy gradient methods), determining which action to take in a given state.
  
    \item \textbf{Value-based (Critic)}: The critic evaluates the actor’s actions by estimating the expected reward, aiding in refining the policy over time.
  
    \item \textbf{Advantage Estimation}: The critic computes the \textit{advantage function} \( A(s, a) = Q(s, a) - V(s) \), which helps the actor decide how much better or worse an action is compared to the expected value of the state.
\end{enumerate}

This synergy allows Actor-Critic algorithms to benefit from both policy optimization (actor) and value estimation (critic), leading to efficient learning in environments with continuous action spaces. The actor component is responsible for exploring and selecting actions based on the current policy, while the critic evaluates these actions by estimating the value function. This dual approach helps mitigate the variance often associated with policy gradient methods, allowing for more stable and reliable updates. Consequently, Actor-Critic algorithms can adaptively improve their policies based on the feedback from the critic, making them particularly well-suited for complex tasks that require nuanced decision-making and planning.

\subsubsection{Deep Deterministic Policy Gradient}
\noindent
Building on the Actor-Critic framework, \nameref{Deep Deterministic Policy Gradient (DDPG)} \citep{lillicrap2015} is an off-policy reinforcement learning algorithm tailored for environments with continuous action spaces. Unlike stochastic policy gradient methods, which involve randomness in action selection, DDPG employs a deterministic policy to map states directly to specific actions. This deterministic approach not only improves sample efficiency but also makes it more effective for tasks that require precise control and fine-tuning, such as robotics and continuous control tasks.
\textbf{Key Components of DDPG}:

\begin{enumerate}
    \item \textbf{Actor Network}: The actor generates actions based on the current state \( s \), using a deterministic policy \( \mu(s|\theta^{\mu}) \), where \( \theta^{\mu} \) are the parameters of the actor network.
    
    \item \textbf{Critic Network}: The critic evaluates the action selected by the actor by estimating the action-value function \( Q(s, a|\theta^{Q}) \), where \( \theta^{Q} \) are the parameters of the critic network.
    
    \item \textbf{Target Networks and Soft Update}: DDPG employs two target networks (one for the actor and one for the critic), denoted as \( \mu' \) and \( Q' \). These target networks are updated gradually using the soft update rule to ensure stable learning and avoid sudden, destabilizing changes to the target values.

    \item \textbf{Ornstein-Uhlenbeck (OU) Action Noise}: DDPG utilizes \textit{OU Action Noise} \citep{hollenstein2022} to encourage exploration in continuous action spaces. The noise is added to the actions selected by the deterministic policy \( \mu(s) \), allowing the agent to explore different parts of the environment by taking slightly varied actions. This is crucial for preventing the agent from getting stuck in local optima.

    \item \textbf{Experience Replay}: DDPG uses a replay buffer to store transitions \( (s_t, a_t, r_t, s_{t+1}) \), which are sampled uniformly to break correlations between consecutive experiences and stabilize training.
    
    \item \textbf{Deterministic Policy}: Unlike stochastic algorithms, DDPG uses a deterministic policy that allows more efficient exploration in continuous action spaces.
\end{enumerate}

\begin{algorithm}[H]
\caption{DDPG Algorithm}
\label{Deep Deterministic Policy Gradient (DDPG)}
\begin{algorithmic}[1]
\State Initialize actor network $\mu(s|\theta^{\mu})$ and critic network $Q(s, a|\theta^Q)$ with parameters $\theta^{\mu}$ and $\theta^Q$
\State Set target networks to initial parameters: $\theta^{Q'} \leftarrow \theta^{Q}$, $\theta^{\mu'} \leftarrow \theta^{\mu}$
\State Initialize replay memory $R$
\For{each episode}
    \State Set a random exploration process $\mathcal{N}$ for action perturbation
    \State Observe and initialize the state $s_0$
    \For{each timestep $t$ in the episode}
        \State Choose action $a_t = \mu(s_t|\theta^{\mu}) + \mathcal{N}_t$
        \State Perform action $a_t$, receive reward $r_t$, and observe new state $s_{t+1}$
        \State Store the experience $(s_t, a_t, r_t, s_{t+1})$ in buffer $R$
        \State Sample a mini-batch of $N$ experiences $(s_i, a_i, r_i, s_{i+1})$ from $R$
        \State Compute target $y_i = r_i + \gamma Q'(s_{i+1}, \mu'(s_{i+1}|\theta^{\mu'})|\theta^{Q'})$
        \State Update the critic by minimizing the loss:
        \begin{equation}
        L = \frac{1}{N} \sum_i \left( y_i - Q(s_i, a_i|\theta^Q) \right)^2
        \end{equation}
        \State Update the actor using the policy gradient:
        \begin{equation}
        \nabla_{\theta^\mu} J \approx \frac{1}{N} \sum_i \nabla_a Q(s, a|\theta^Q) \nabla_{\theta^\mu} \mu(s|\theta^\mu)
        \end{equation}
        \State Adjust target networks by soft updates:
        \begin{equation}
        \theta^{Q'} \leftarrow \tau \theta^Q + (1 - \tau) \theta^{Q'}
        \end{equation}
        \begin{equation}
        \theta^{\mu'} \leftarrow \tau \theta^\mu + (1 - \tau) \theta^{\mu'}
        \end{equation}
    \EndFor
\EndFor
\end{algorithmic}
\end{algorithm}

\subsubsection{Twin Delayed Deep Deterministic Policy Gradient}
This algorithm \citep{fujimoto2018} is an enhancement of the DDPG algorithm, specifically designed to address the instability and overestimation issues commonly found in DDPG. By incorporating techniques such as target network updates and noise clipping, TD3 introduces several key improvements that significantly increase the stability and performance of the learning process in continuous action spaces.

% \vspace{0.5em}
\textbf{Key Improvements Over DDPG:}

\begin{enumerate}
    \item \textbf{Clipped Double Q-Learning}: 
    In DDPG, the critic network tends to overestimate Q-values, which can lead to suboptimal policies. TD3 addresses this by using two critic networks, denoted as \( Q'_1 \) and \( Q'_2 \), and taking the minimum of the two Q-values to calculate the target:
    \[
    y_i = r_i + \gamma \min_{j=1,2} Q'_j(s_{i+1}, \mu'(s_{i+1}|\theta^{\mu'})) 
    \]
    This reduction in overestimation bias improves the stability of the learning process.

    \item \textbf{Delayed Policy Updates}: 
    TD3 updates the actor network (policy) less frequently than the critic networks. For every \( d \) updates to the critic, the actor is updated once. This delay allows the critic to better estimate the Q-values before updating the policy, reducing the likelihood of harmful updates:
    \[
    \theta^{\mu} \leftarrow \theta^{\mu} + \alpha \nabla_{\theta^{\mu}} J(\theta^{\mu})
    \]
    where \( \alpha \) is the learning rate of the actor network.

    \item \textbf{Target Policy Smoothing}: 
    To prevent the policy from exploiting narrow peaks in Q-values, TD3 adds noise to the target policy during critic updates. This noise is clipped to ensure it remains within a reasonable range, which results in smoother policy updates:
    \[
    a' = \mu'(s_{i+1}|\theta^{\mu'}) + \epsilon, \quad \epsilon \sim \text{clip}(\mathcal{N}(0, \sigma), -c, c)
    \]
    where \( \epsilon \) is the added noise, and \( c \) is a constant used for clipping. This encourages more stable exploration and smoother learning.\\
\end{enumerate}

These enhancements make TD3 more robust in handling continuous action spaces and provide significantly better performance compared to DDPG, particularly in environments that require fine control, such as autonomous navigation and robotics. The improvements in TD3, including the addition of noise clipping and a more frequent update mechanism for the target networks, address the challenges of overestimation bias and instability often encountered in reinforcement learning. Hyperparameters that are modified in TD3 are detailed in \cref{tab:combined_hyperparameters}. Notably, TD3 allows for a greater learning rate for both the Actor and Critic Networks, which contributes to more efficient learning and faster convergence in complex tasks.
\begin{table}[h]
    \captionsetup{justification=centering, font=small, labelsep=period, skip=0.5em} % Adjust caption properties
    \caption{Comparison of Hyperparameters - DDPG vs. TD3}
    \label{tab:combined_hyperparameters}
    \centering
    \small % Reduces font size to fit better in the column
    \renewcommand{\arraystretch}{1.2} % Adjust row height
    \begin{tabular}{|p{3.6cm}|p{7.2cm}|p{2.2cm}|p{2.2cm}|} \hline
        \textbf{Hyperparameter} & \textbf{Description} & \textbf{DDPG Value} & \textbf{TD3 Value} \\ \hline
        Alpha & Learning rate for the actor network determines the size of each update step & 0.0001 & 0.0005 \\ \hline
        Beta & Learning rate for the critic network determines the size of each update step & 0.001 & 0.005 \\ \hline
        Tau & Controls the update rate of the target network weights & 0.001 & 0.001 \\ \hline
        Replay Buffer Size & Maximum capacity of stored experiences & 100,000 & 100,000 \\ \hline
        Batch Size & Number of samples drawn per training update & 64 & 64 \\ \hline
        Noise Clip & Clips the noise to this range to stabilize exploration & -- & 0.5 \\ \hline
        Policy Update Frequency & Number of steps to delay policy updates & -- & 2 \\ \hline
    \end{tabular}
\end{table}

\begin{figure}[h]
    \centering
    % First subfigure
    \begin{subfigure}[t]{0.32\textwidth}
        % {16A}
        \centering
        \fbox{\includegraphics[width=5.3cm, height=4.4cm]{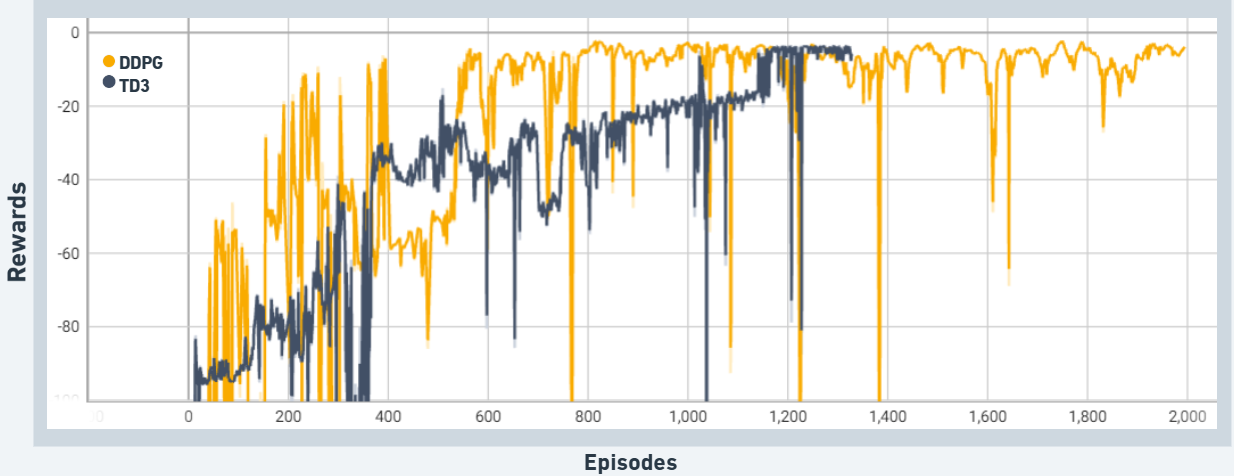}}
        \caption{Rewards collected vs episode for Scenario 1}
        \label{fig:Rewards Scenario 1}
    \end{subfigure}
    \hfill
    % Second subfigure
    \begin{subfigure}[t]{0.32\textwidth}
        % {16B}
        \centering
        \fbox{\includegraphics[width=5.3cm, height=4.4cm]{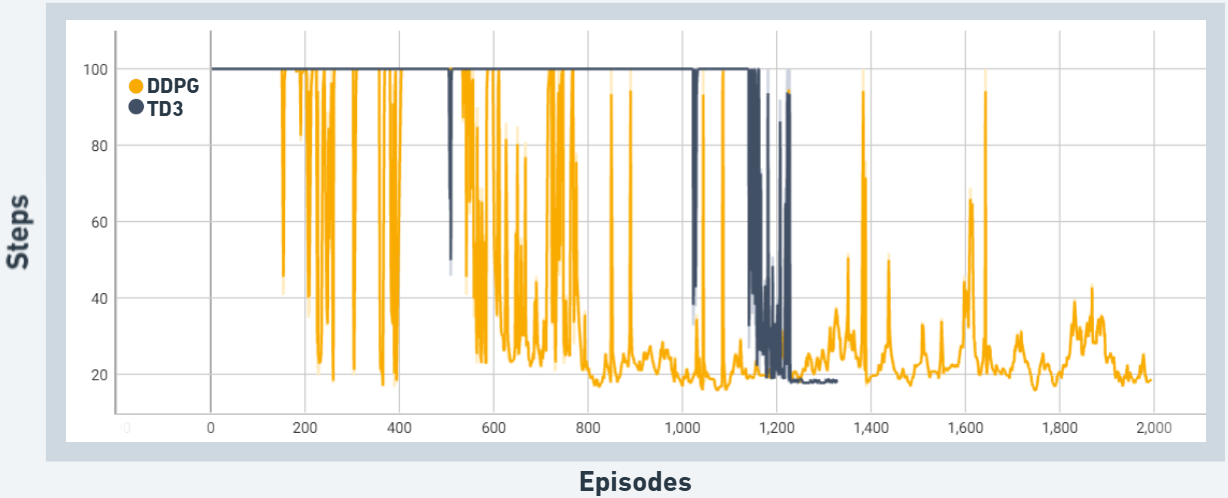}}
        \caption{Steps taken vs episode for Scenario 1}
        \label{fig:Steps Scenario 1}
    \end{subfigure}
    \hfill
    % Third subfigure
    \begin{subfigure}[t]{0.32\textwidth}
        % {16C}
        \centering
        \fbox{\includegraphics[width=5.3cm, height=4.4cm]{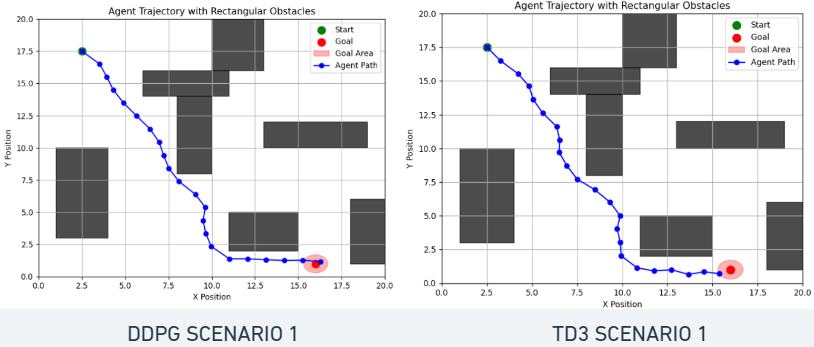}}
        \caption{Planned path for DDPG \& TD3 for Scenario 1}
        \label{fig:Planned path Scenario 1}
    \end{subfigure}

    \begin{subfigure}[t]{0.32\textwidth}
        % {17A}
        \centering
        \fbox{\includegraphics[width=5.3cm, height=4.4cm]{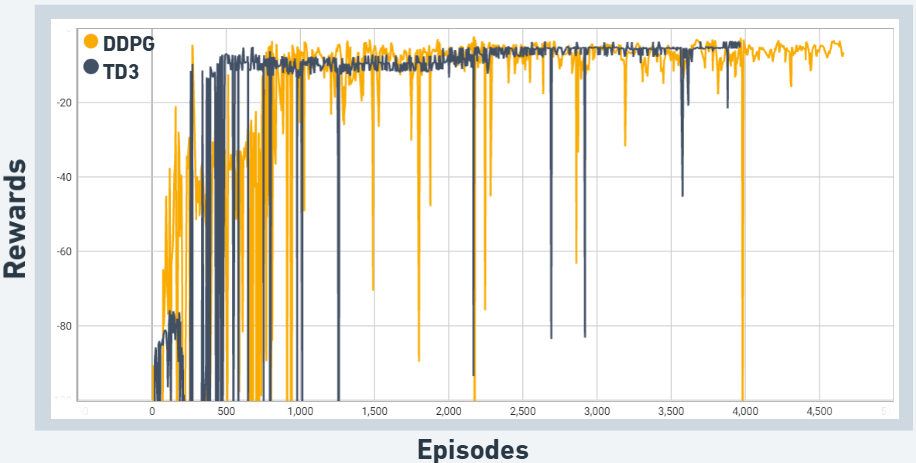}}
        \caption{Rewards collected vs episode for Scenario 2}
        \label{fig:Rewards Scenario 2}
    \end{subfigure}
    \hfill
    % Second subfigure
    \begin{subfigure}[t]{0.32\textwidth}
        % {17B}
        \centering
        \fbox{\includegraphics[width=5.3cm, height=4.4cm]{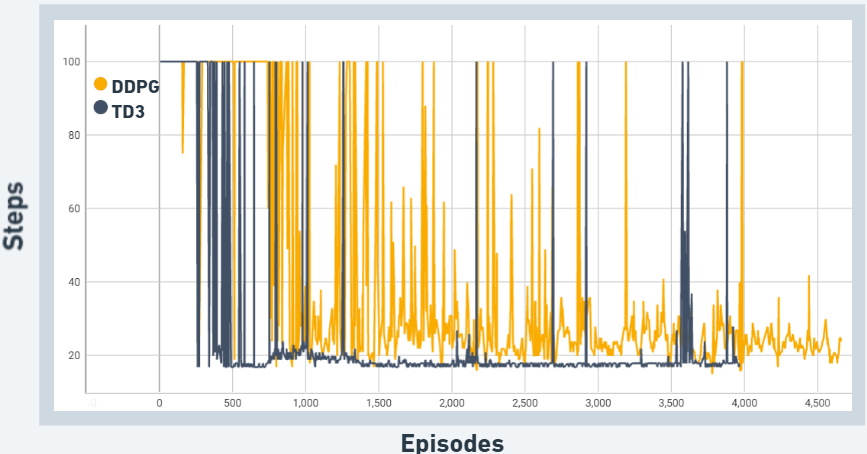}}
        \caption{Steps taken vs episodes for Scenario 2}
        \label{fig:Steps Scenario 2}
    \end{subfigure}
    \hfill
    % Third subfigure
    \begin{subfigure}[t]{0.32\textwidth}
        % {17C}
        \centering
        \fbox{\includegraphics[width=5.3cm, height=4.4cm]{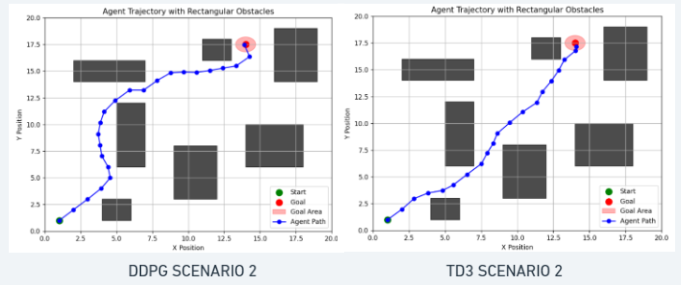}}
        \caption{Planned Path for Scenario 2}
        \label{fig:Planned Path Scenario 2}
    \end{subfigure}

    \begin{subfigure}[t]{0.32\textwidth}
        % {18A}
        \centering
        \fbox{\includegraphics[width=5.3cm, height=4.6cm]{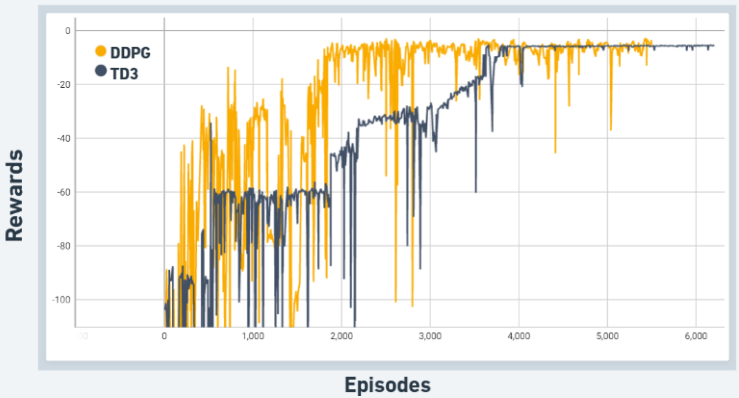}}
        \caption{Rewards collected vs episode for Scenario 3}
        \label{fig:Rewards Scenario 3}
    \end{subfigure}
    \hfill
    % Second subfigure
    \begin{subfigure}[t]{0.32\textwidth}
        % {18B}
        \centering
        \fbox{\includegraphics[width=5.3cm, height=4.6cm]{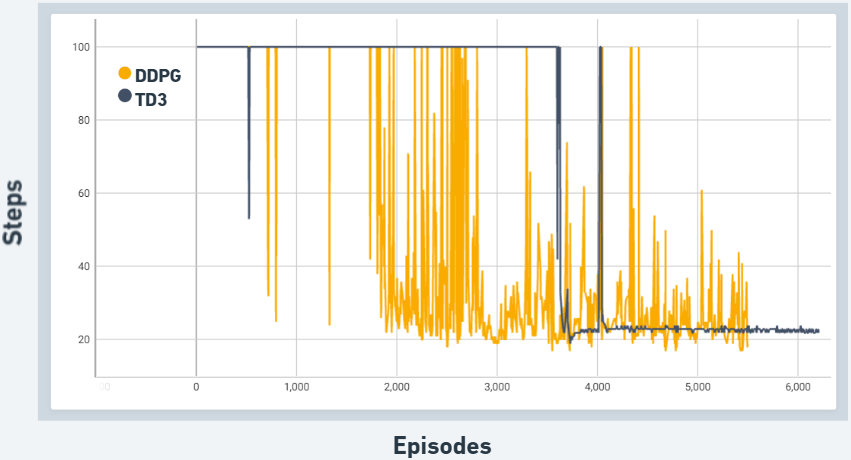}}
        \caption{Steps taken vs episode for Scenario 3}
        \label{fig:Steps Scenario 3}
    \end{subfigure}
    \hfill
    % Third subfigure
    \begin{subfigure}[t]{0.32\textwidth}
        % {18C}
        \centering
        \fbox{\includegraphics[width=5.3cm, height=4.6cm]{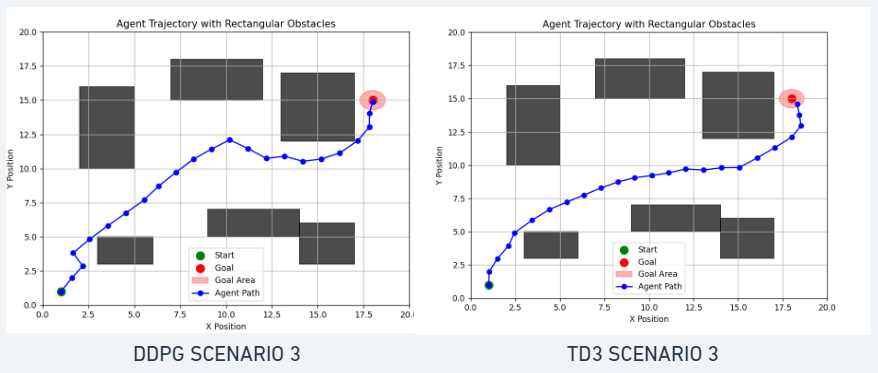}}
        \caption{Planned Path for Scenario 3}
        \label{fig:Planned Path Scenario 3}
    \end{subfigure}
    \caption{Training performance of TD3 \& DDPG across three scenarios}
    \label{fig:Scenario 3 Results}
\end{figure}

\subsection{Observations}
The successful implementation of DDPG and TD3 allows for training both algorithms within the 2D Continuous environment, facilitating a comparative analysis of their performance.

% \vspace{0.5em}
\textbf{Convergence Criteria}: The convergence is defined as the average reward collected per episode for the last trailing 100 episodes being less than -5, and the trailing average steps taken to reach the goal area being less than 30 steps.

*\textit{In this experiment, stability is being quantified as the ability of the agent to be able to minimize the number of steps each episode after it has started converging.}

In Scenario 1, as illustrated in \cref{fig:Rewards Scenario 1}, DDPG began to converge early, around the 800th episode, primarily through its random exploration strategies. However, stability was not achieved until approximately the 2000th episode, indicating that while DDPG could identify rewarding actions, it struggled to maintain consistent performance during this period (see \cref{fig:Steps Scenario 1}). In contrast, TD3 started converging later, around the 1100th episode, but exhibited a faster and more stable convergence process within the following 200 episodes. This suggests that TD3's enhancements effectively mitigated the instability issues seen in DDPG, allowing for more reliable learning. The paths planned by both TD3 and DDPG during this scenario can be observed in \cref{fig:Planned path Scenario 1}, highlighting their differing strategies for navigation and goal achievement.

In Scenario 2, as seen in \cref{fig:Rewards Scenario 2}, TD3 again exhibited earlier convergence compared to DDPG. Although the difference was not as pronounced as in Scenario 1, it is noteworthy that TD3 maintained greater stability post-convergence, indicating its robustness in adapting to the task. This improved stability allowed TD3 to consistently achieve high rewards over multiple episodes, suggesting a more reliable learning process. Furthermore, the planned path generated by TD3 was observed to be more efficient and shorter than that produced by DDPG (see \cref{fig:Planned Path Scenario 2}), reflecting TD3's ability to optimize its navigation strategy effectively. TD3 also outperformed DDPG in terms of stability, as observed in \cref{fig:Steps Scenario 2}, showcasing its advantages in maintaining performance even in challenging conditions.

In Scenario 3, as shown in \cref{fig:Rewards Scenario 3}, an inverse trend was observed: DDPG converged earlier, around the 6000th episode, while TD3 approached convergence after approximately 4000 episodes. However, TD3's average trailing rewards did not exceed -5, indicating incomplete convergence. Despite this, \cref{fig:Steps Scenario 3} reveals that TD3 maintained excellent stability after initiating convergence, whereas DDPG exhibited significant inconsistency. Furthermore, the path planned by TD3 was more efficient and straightforward compared to that of DDPG (\cref{fig:Planned Path Scenario 3}).

\subsection{Results and Analysis}

The results presented in \cref{tab:performance_comparison_scenarios} are obtained by averaging each metric across five training trials for robustness. The \textbf{stability measure} is calculated by setting a threshold for the number of steps and finding the fraction of episodes after convergence starts in which the agent requires fewer steps than the threshold to solve the environment. The formula for the stability measure is:

\[
\text{Stability Measure} = \frac{\text{No. of Eps with Steps below threshold}}{\text{Total Eps After Convergence Starts}}
\]

Key insights from the results include:

\begin{enumerate}
    \item In \textbf{Scenario 1} and \textbf{Scenario 2}, TD3 starts converging later than DDPG but converges faster thereafter, indicating a slower warm-up but quicker learning rate for TD3.
    \item In \textbf{Scenario 3}, TD3 plans a viable path but does not achieve full convergence, whereas DDPG does, albeit with more steps.
    \item TD3 shows significant improvement in the \textbf{stability measure}, maintaining a consistent number of steps to solve the environment, with a \textbf{19.90\% average stability improvement} over DDPG.
    \item Regarding \textbf{training time}, TD3 generally completes training faster across most scenarios, except Scenario 3, where it takes longer due to environmental complexities. The overall \textbf{training time improvement is 5.96\%}.
    \item TD3 also demonstrates improvements in path length compared to DDPG, with a \textbf{13.72\% average path length improvement}, which becomes crucial in larger environments, such as agricultural fields.
\end{enumerate}

\begin{table}[h]
    \caption{Comparison of TD3 and DDPG across three scenarios}
    \label{tab:performance_comparison_scenarios}
    \centering
    \renewcommand{\arraystretch}{1.4} % Adjust row height for better spacing
    \setlength{\tabcolsep}{6pt} % Adjust cell padding for better spacing
    \begin{tabular}{|c|c|c|c|c|c|c|}
        \toprule
        \multirow{2}{*}{\textbf{Performance Metric}} & \multicolumn{2}{c|}{\textbf{Scenario 1}} & \multicolumn{2}{c|}{\textbf{Scenario 2}} & \multicolumn{2}{c|}{\textbf{Scenario 3}} \\
        \cmidrule(lr){2-3} \cmidrule(lr){4-5} \cmidrule(lr){6-7}
        & \textbf{TD3} & \textbf{DDPG} & \textbf{TD3} & \textbf{DDPG} & \textbf{TD3} & \textbf{DDPG} \\
        \midrule
        Convergence starts & 1200 & 800 & 780 & 340 & 3660 & 1880 \\
        Completely converged & 1327 & 1994 & 3962 & 4659 & - & 5500 \\
        Path length & 20.02 & 24.23 & 21.60 & 24.91 & 22.41 & 25.05 \\
        Stability Measure & 95.05\% & 85.08\% & 96.65\% & 75.03\% & 96.24\% & 80.11\% \\
        Training time (mins) & 18 & 25 & 15 & 20 & 30 & 22 \\
        \bottomrule
    \end{tabular}
\end{table}
\textit{*Training time has been rounded off.}

In summary, TD3’s stability, faster convergence, and efficient path planning (in Scenarios 1 and 2) make it well-suited for 3D simulation environments. Its consistent performance and reduced training time are advantageous for complex, dynamic settings like agricultural fields, where stable navigation and adaptability are essential.

\section{Experimentation - Gazebo Env with ROS}

\subsection{3D Agricultural Environment Design}
The 3D agricultural environment used in this research was carefully designed to simulate a real-world farming field, incorporating common challenges faced by Unmanned Ground Vehicles (UGVs) in precision agriculture. The environment was built within the Gazebo simulator, which provides a physics-based platform for testing and validating autonomous navigation strategies. In this environment, the UGV must navigate autonomously, avoid obstacles, and reach dynamically placed goals, mimicking the tasks a UGV would perform in an actual agricultural setting. The dimensions of the robot and the agricultural environment are mentioned in \cref{table:dimensions}.

\begin{table}[h]
    \captionsetup{justification=centering, font=small, labelsep=period, skip=0.5em} % Adjust caption properties
    \caption{Dimensions of the Robot and Environment}
    \label{table:dimensions}
    \centering
    \small % Reduces font size to fit better in the column
    \renewcommand{\arraystretch}{1.2} % Adjust row height
    \begin{tabular}{|>{\centering\arraybackslash}p{3.8cm}|>{\centering\arraybackslash}p{3.5cm}|} \hline 
        \textbf{Specification} & \textbf{Dimensions} \\ \hline 
        Robot Dimensions & 14 cm x 14 cm x 14 cm \\ \hline 
        Environment Dimensions & 10 m x 8 m \\ \hline 
    \end{tabular}
\end{table}

\begin{itemize}
    \item \textbf{Environment Layout and Terrain}: 
    The 3D environment mimics an agricultural field with structured and unstructured elements, featuring:
    terrain and boundaries (e.g., fences) enclose the environment, ensuring the UGV operates within a designated area, akin to real farm limits.
    
    \item \textbf{Obstacles and Hazards}:
Static obstacles, including crops and stones, are modeled in Blender to create a realistic and cluttered layout that challenges the UGV's path-planning capabilities. Additionally, a dynamic obstacle, represented by a dummy human, is introduced to increase environmental complexity. This dynamic obstacle serves as a proxy for real-life agricultural challenges, such as moving livestock and humans, further simulating practical scenarios in precision agriculture.
\end{itemize}

\begin{figure}[h]
    \centering
    % First subfigure
    \begin{subfigure}[t]{0.32\textwidth}
        % {12A}
        \centering
        \fbox{\includegraphics[width=5cm, height=3cm]{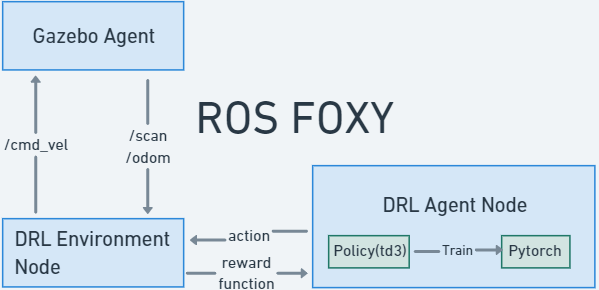}}
        \caption{ROS-Gazebo Pipeline}
        \label{fig:ROS-Gazebo Pipeline}
    \end{subfigure}
    % Second subfigure
    \begin{subfigure}[t]{0.32\textwidth}
        % {12B}
        \centering
        \fbox{\includegraphics[width=4cm, height=4cm]{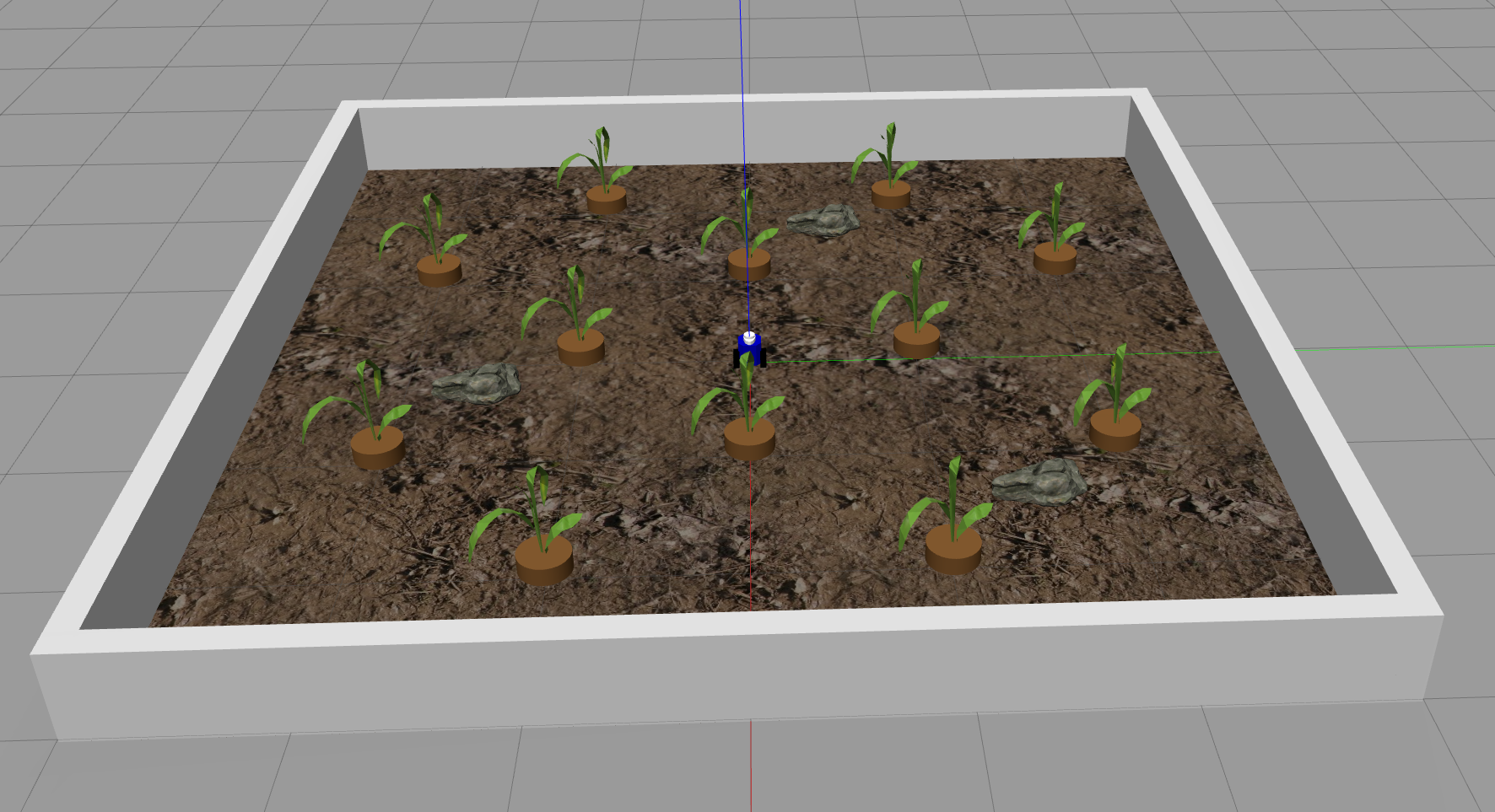}}
         \caption{Environment Layout-Static}
        \label{fig:Environment Layout}
    \end{subfigure}
    \begin{subfigure}[t]{0.32\textwidth}
        % {12C}
        \centering
    \fbox{\includegraphics[width=4cm, height=4cm]{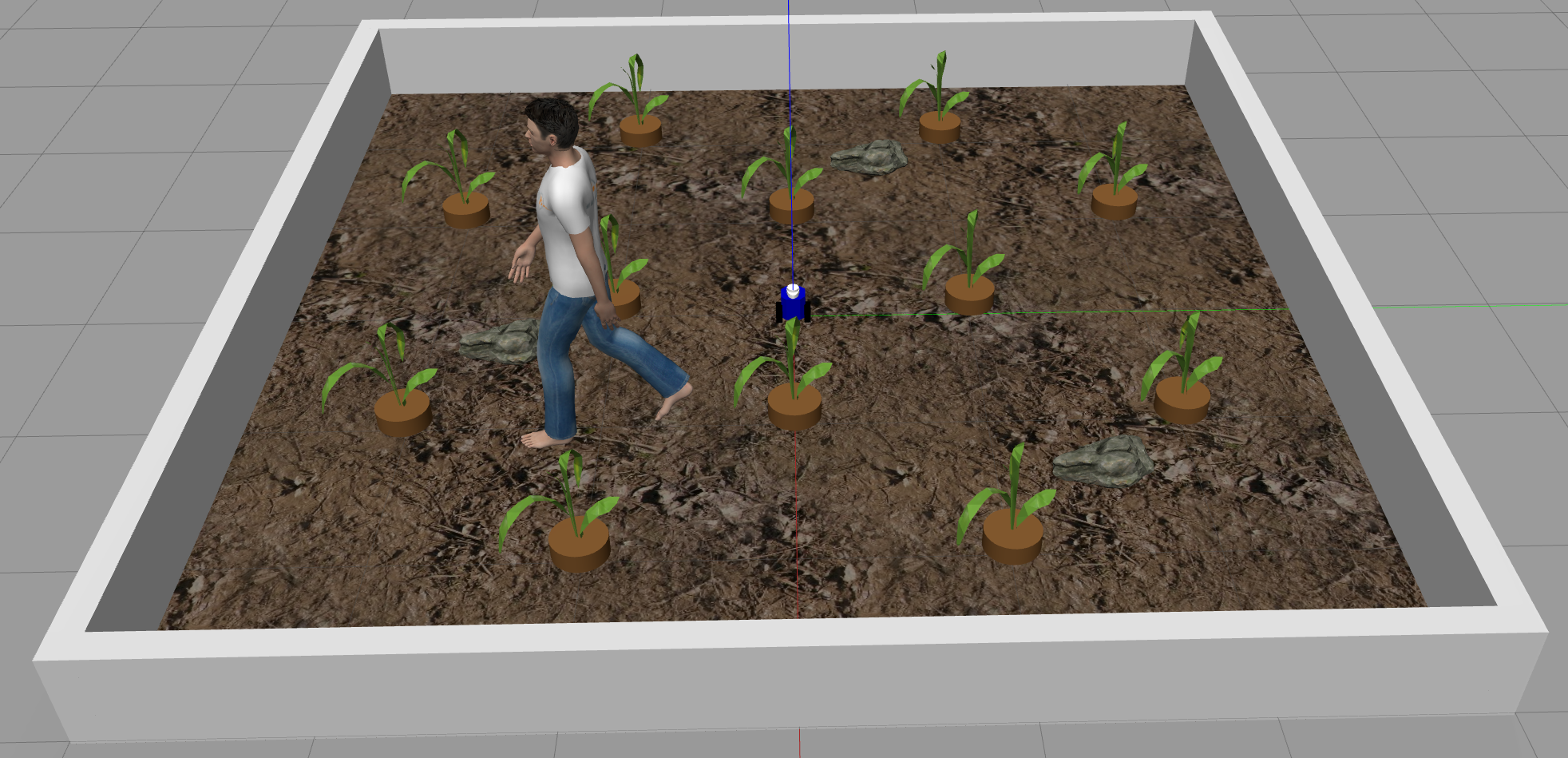}}
        \caption{Environment Layout-Dynamic}
        \label{fig:Environment Layout}
    \end{subfigure}
    \caption{3D Environment Structure \& Pipeline}
    \label{fig:d3qn-comparison}
\end{figure}

\subsection{DRL Model Integration with ROS}

% \vspace{0.8em}
\subsubsection{System Architecture}
The training architecture comprises three main components, illustrated in the diagram \cref{fig:ROS-Gazebo Pipeline}:

\begin{itemize}
    \item \textbf{Gazebo/Physical Robot}
    The Gazebo simulator offers a realistic 3D environment for the UGV, simulating sensors like LiDAR (/scan) and odometry (/odom). In real-world applications, this is replaced by a physical robot with identical sensors. The UGV receives movement commands for linear and angular velocities through the /cmd\_vel topic in ROS 2.
    
    \item \textbf{DRL Environment Node}
    This node acts as a bridge between the environment and the DRL agent, managing the flow of sensor data (e.g., obstacle distances) to the agent. It also relays actions (velocity and steering commands) from the DRL agent to the UGV and computes reward signals based on metrics (e.g., obstacle proximity, goal distance).
    
    \item \textbf{DRL Agent Node}
    Responsible for decision-making, this node contains the policy network trained with the TD3 algorithm. It determines actions (linear and angular velocities) based on observations from the environment. PyTorch facilitates training, where the agent maximizes cumulative rewards by refining its navigation policy. The Train() function updates model parameters based on environmental feedback.
\end{itemize}

\subsubsection{System Workflow}
The system workflow is divided into the following three sections:

\begin{enumerate}
    \item \textbf{Training Pipeline for Navigation System}: 
    The DRL-based navigation system’s training pipeline integrates various components for the Unmanned Ground Vehicle (UGV) to learn efficient path-planning in a 3D environment. Utilizing the Gazebo simulator or a physical robot, it communicates with the Deep Reinforcement Learning (DRL) agent via ROS 2, allowing the UGV to discover optimal policies through trial and error, reinforced by reward feedback.
    
    \begin{itemize}
    \item \textbf{Observation and State Representation}:
    At each time step, the DRL environment node gathers sensor data (LiDAR, odometry) to provide the UGV's current state (position, orientation, obstacle proximity) to the DRL agent node, forming the observation space for decision-making.
    
    \item \textbf{Action Selection}:
    Based on observations, the DRL agent selects an action (linear and angular velocities) using the trained policy, which is sent to the DRL environment node to control the UGV in Gazebo or the physical robot via the /cmd\_vel ROS 2 topic.
    
    \item \textbf{Reward Calculation}:
    In the 3D environment, the reward function is more complex than the 2D, taking into account both the agent’s linear and angular movements. The reward structure is defined in \cref{tab:reward_structure}.
    
    \item \textbf{Policy Update (Training)}:
    Using the reward and new observation, the DRL agent updates its policy via reinforcement learning (e.g., TD3), aiming to maximize cumulative rewards by improving decision-making.
    
    \item \textbf{Iteration and Learning}:
    This cycle repeats across multiple episodes, with the UGV exploring the environment and enhancing its navigation strategy. Over time, the DRL agent minimizes collisions, optimizes path efficiency, and accelerates goal achievement.
    
    \item \textbf{Dynamic Goal Adaptation}:
    To prevent overfitting and encourage exploration, the position of the goal is randomly changed at the beginning of each episode. This approach ensures the agent does not learn only a single path but instead generalizes across varying scenarios.
    
    \item \textbf{Moving Obstacles}:
   The model trained in the static environment was retrained in the dynamic environment using transfer learning, introducing a moving obstacle (dummy human) to increase complexity and introduce external disturbance. The telemetry of the human is designed such that it will only collide with the agent and not the crops.
\end{itemize}

    This training pipeline enables the DRL agent to effectively navigate complex agricultural fields, avoid obstacles, and reach dynamic goals, continually enhancing performance through experience and feedback.

\begin{table}[h]
    \captionsetup{justification=centering, font=small, labelsep=period, skip=0.5em} % Adjust caption properties
    \caption{Reward Structure in 3D Environment}
    \label{tab:reward_structure}
    \centering
    \small % Reduces font size to fit better in the column
    \begin{tabular}{|p{4cm}|p{6.5cm}|p{5.0cm}|} \hline 
    
        \textbf{Component} & \textbf{Reward/Penalty Formula} & \textbf{Description} \\ \hline 
        Angular Deviation Penalty & 
        $r_{yaw} = -1 \times \left| goal\_angle \right|$ & 
        Penalizes deviation of the agent's orientation (yaw angle) from the goal direction. \\ \hline 

        Angular Velocity Penalty & 
        $r_{vangular} = -1 \times (action\_angular^2)$ & 
        Penalizes excessive angular motion based on squared angular velocity. \\ \hline 

        Distance-Based Reward & 
        $r_{distance} = \frac{2 \times goal\_dist\_initial}{goal\_dist\_initial + goal\_dist} - 1$ & 
        Rewards the agent for minimizing the distance to the goal. \\ \hline 

        Obstacle Proximity Penalty & 
        $r_{obstacle} = 
        \begin{cases} 
        -20 & \text{if } min\_obstacle\_dist < 0.22 \\ 
        0 & \text{otherwise} 
        \end{cases}$ & 
        Penalizes the agent for moving too close to obstacles (within 0.22 units). \\ \hline 

        Linear Velocity Penalty & 
        $r_{vlinear} = -1 \times \left( (0.22 - action\_linear) \times 10 \right)^2$ & 
        Penalizes deviation of the agent's linear velocity from a safe range. \\ \hline 

        Constant Step Penalty & 
        $-1$ & 
        Penalizes every step to encourage faster goal-reaching behavior. \\ \hline 

        Success Reward & 
        $+2500$ & 
        Reward for successfully reaching the goal. \\ \hline 

        Failure Penalty & 
        $-2000$ & 
        Penalty for colliding with obstacles or the environment’s boundary. \\ \hline 
        
    \end{tabular}
\end{table}

\end{enumerate}

\subsection{Observations}
The TD3 algorithm was run over multiple episodes, allowing the UGV to learn navigation using the defined reward structure. Performance was tracked through average rewards and steps per episode.

\begin{figure}[h]
    \centering
    % First subfigure
    \begin{subfigure}[t]{0.45\textwidth}
        % {21A}
        \centering
        \fbox{\includegraphics[width=\linewidth, height=4.8cm]{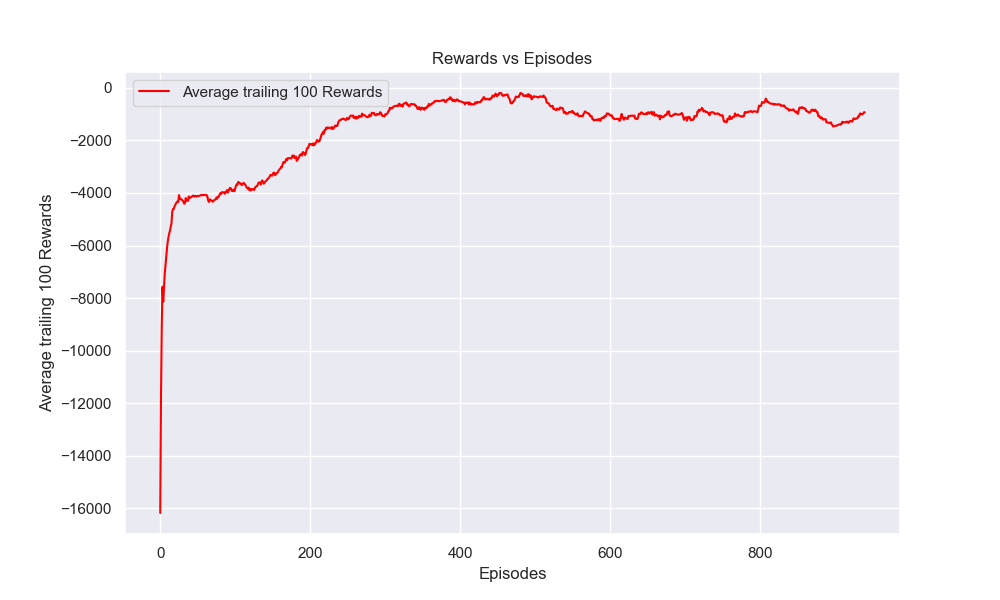}}
        \caption{Reward Graph}
        \label{fig:Reward Static Graph}
    \end{subfigure}
    \hfill
    % Second subfigure
    \begin{subfigure}[t]{0.45\textwidth}
        % {21B}
        \centering
        \fbox{\includegraphics[width=\linewidth, height=4.8cm]{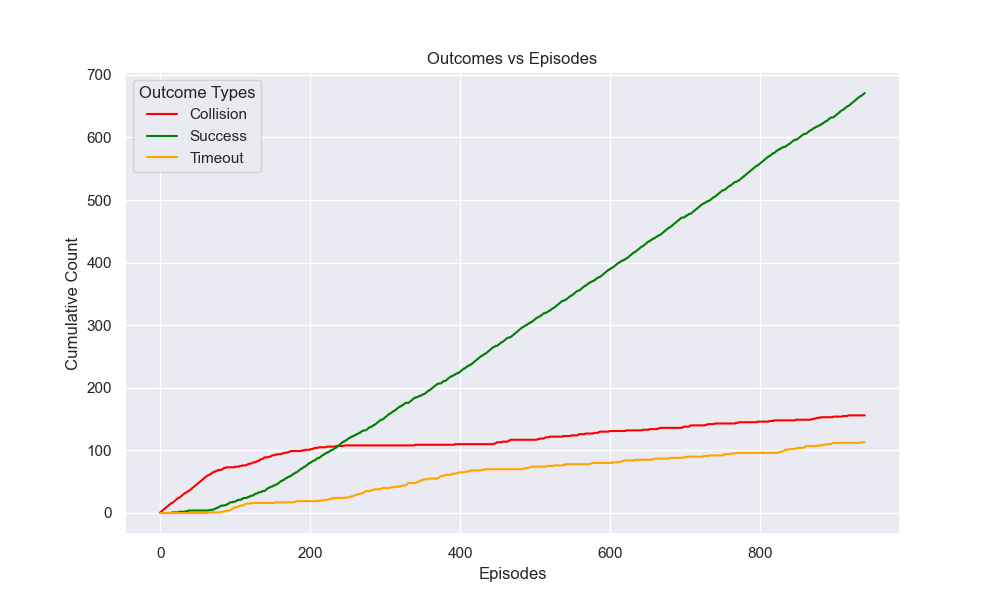}}
        \caption{Episode Outcomes}
        \label{fig:Static Episode Outcomes}
    \end{subfigure}
    \caption{Training performance of TD3 in 3D Static Environment}
    \label{fig:reward-episode-outcomes}
\end{figure}

\begin{figure}[h]
    \centering
    % First subfigure
    \begin{subfigure}[t]{0.45\textwidth}
        % {21A}
        \centering
        \fbox{\includegraphics[width=\linewidth, height=4.8cm]{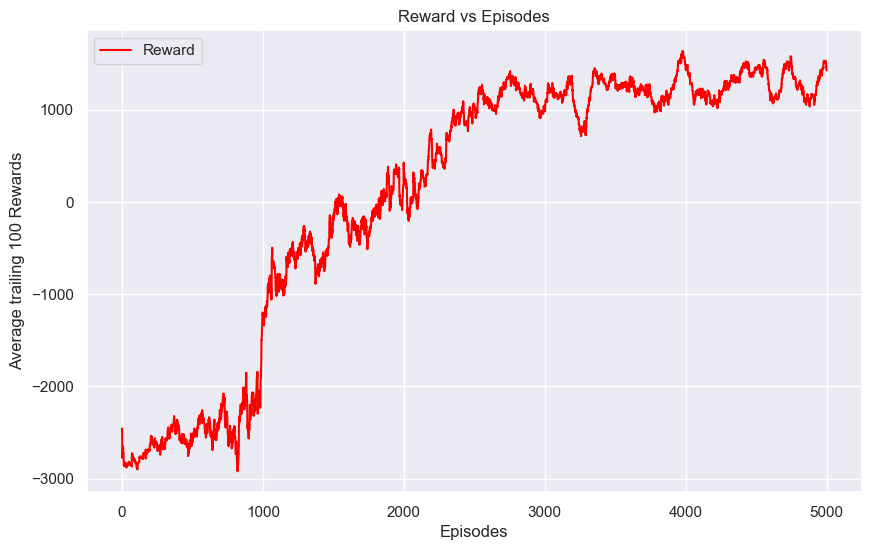}}
        \caption{Reward Graph}
        \label{fig:Dynamic Reward Graph}
    \end{subfigure}
    \hfill
    % Second subfigure
    \begin{subfigure}[t]{0.45\textwidth}
        % {21B}
        \centering
        \fbox{\includegraphics[width=\linewidth, height=4.8cm]{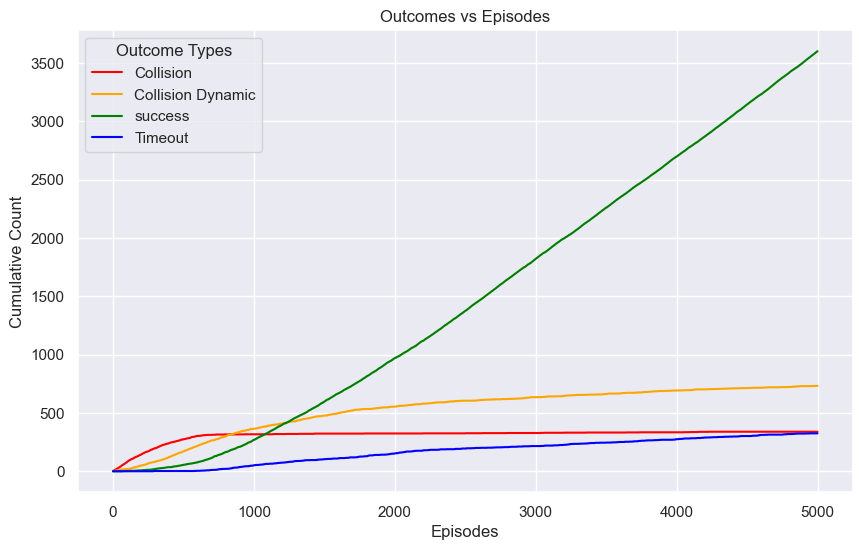}}
        \caption{Episode Outcomes}
        \label{fig:Dynamic Episode Outcomes}
    \end{subfigure}
    \caption{Training performance of TD3 in 3D Dynamic Environment}
    \label{fig:reward-episode-outcomes}
\end{figure}

\begin{itemize}
    \item \textbf{Cumulative Rewards}:  
    The cumulative rewards collected by the UGV during static training are presented in \cref{fig:Reward Static Graph}. Initial fluctuations reflect the UGV’s exploration phase as it learns to navigate obstacles and reach dynamically placed goals. Over time, the steady increase in average rewards indicates that the UGV is developing an effective navigation policy. In contrast to the smooth reward trajectory in the static environment, the environment with dynamic obstacles exhibits a more irregular trend (\cref{fig:Dynamic Reward Graph}). However, by the 2500th episode, the agent is able to accumulate rewards of approximately 1000.

    \item \textbf{Successful Outcomes}:  
    A steady increase in successful outcomes is observed in both static and dynamic environments (\cref{fig:Static Episode Outcomes} \& \cref{fig:Dynamic Episode Outcomes}). While the static environment minimizes collisions by the 700th episode, Timeouts increase due to the agent's strategy of waiting for the next goal to spawn when a goal is placed too close to the crops, ensuring no harm to the crops. Training is stopped at the 900th episode to facilitate the transfer of learning to the dynamic environment.

    \item \textbf{Dynamic Environment Training}:  
    The dynamic environment demonstrates notable training efficiency, benefiting from the pre-trained static agent. Collisions with both dynamic and static obstacles stabilize by the 1700th episode, while successful outcomes increase rapidly. Training is continued until the 5000th episode to further enhance performance.
\end{itemize}

\begin{figure}[h]
    \centering
    % First subfigure
    \begin{subfigure}[t]{0.24\textwidth}
        \centering
        \fbox{\includegraphics[width=\linewidth, height=4cm]{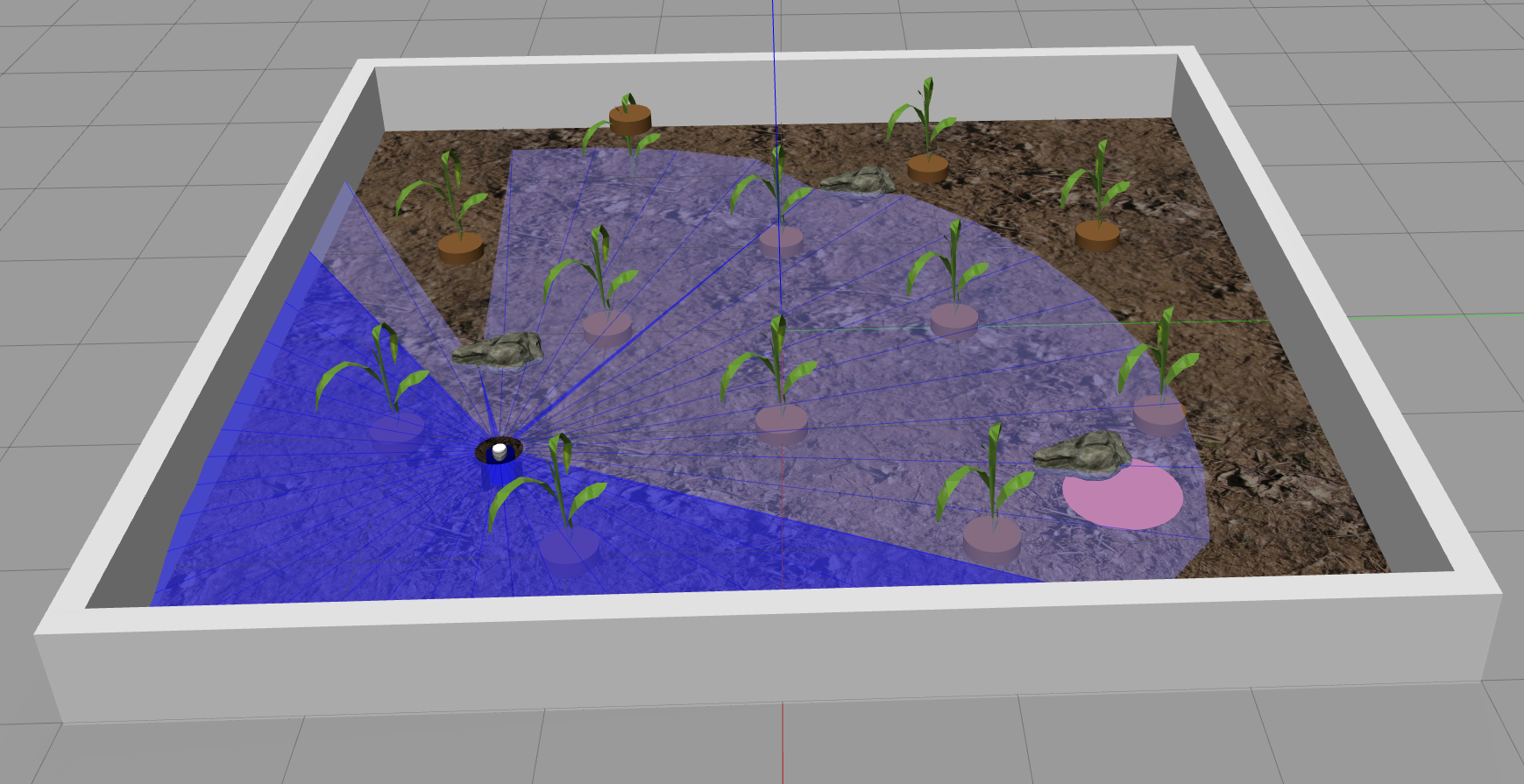}}
        \caption{Goal 1}
        \label{fig:Goal 1}
    \end{subfigure}
    % \hfill
    % Second subfigure
    \begin{subfigure}[t]{0.24\textwidth}
        \centering
        \fbox{\includegraphics[width=\linewidth, height=4cm]{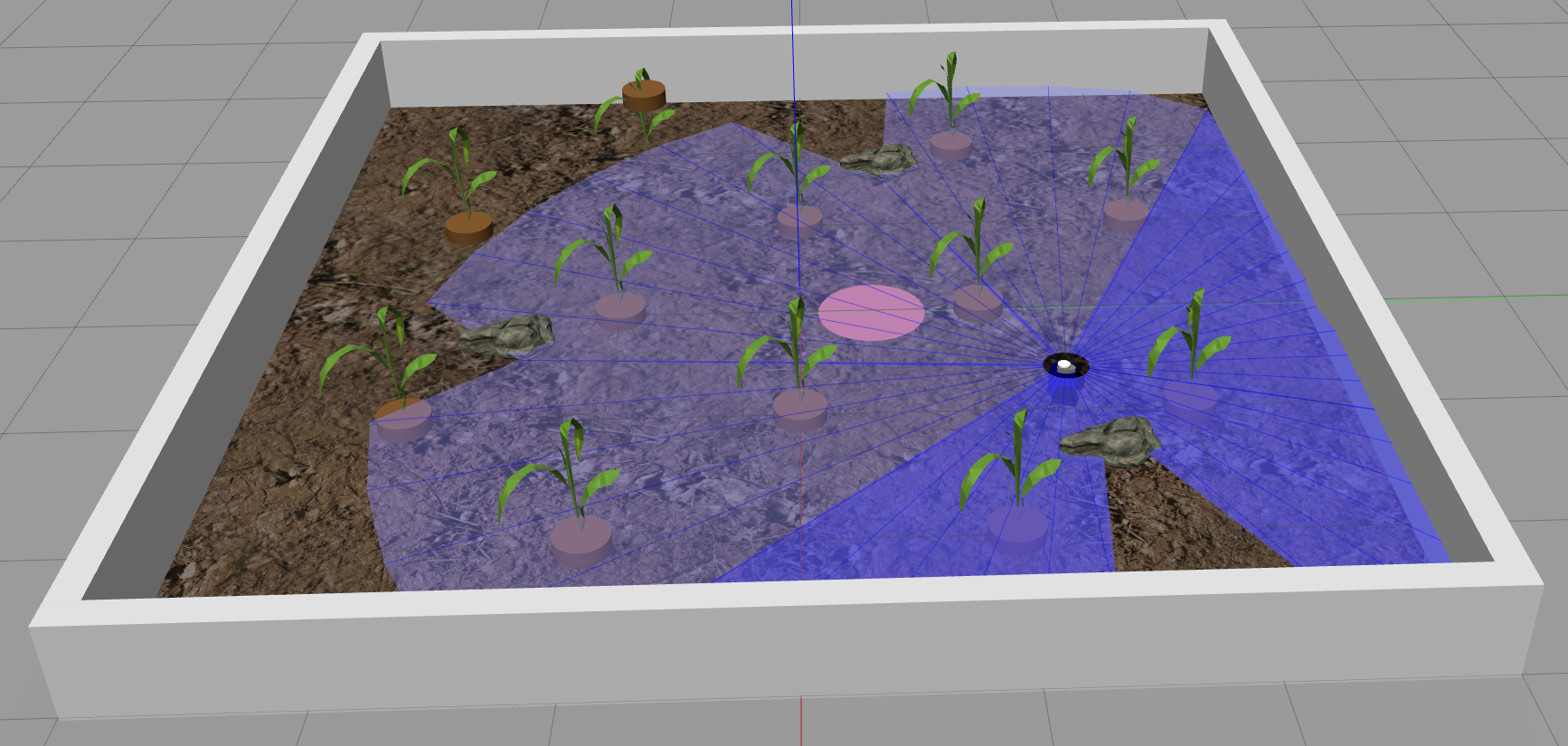}}
        \caption{Goal 2}
        \label{fig:Goal 2}
    \end{subfigure}
    % \hfill
    % Third subfigure
    \begin{subfigure}[t]{0.24\textwidth}
        \centering
        \fbox{\includegraphics[width=\linewidth, height=4cm]{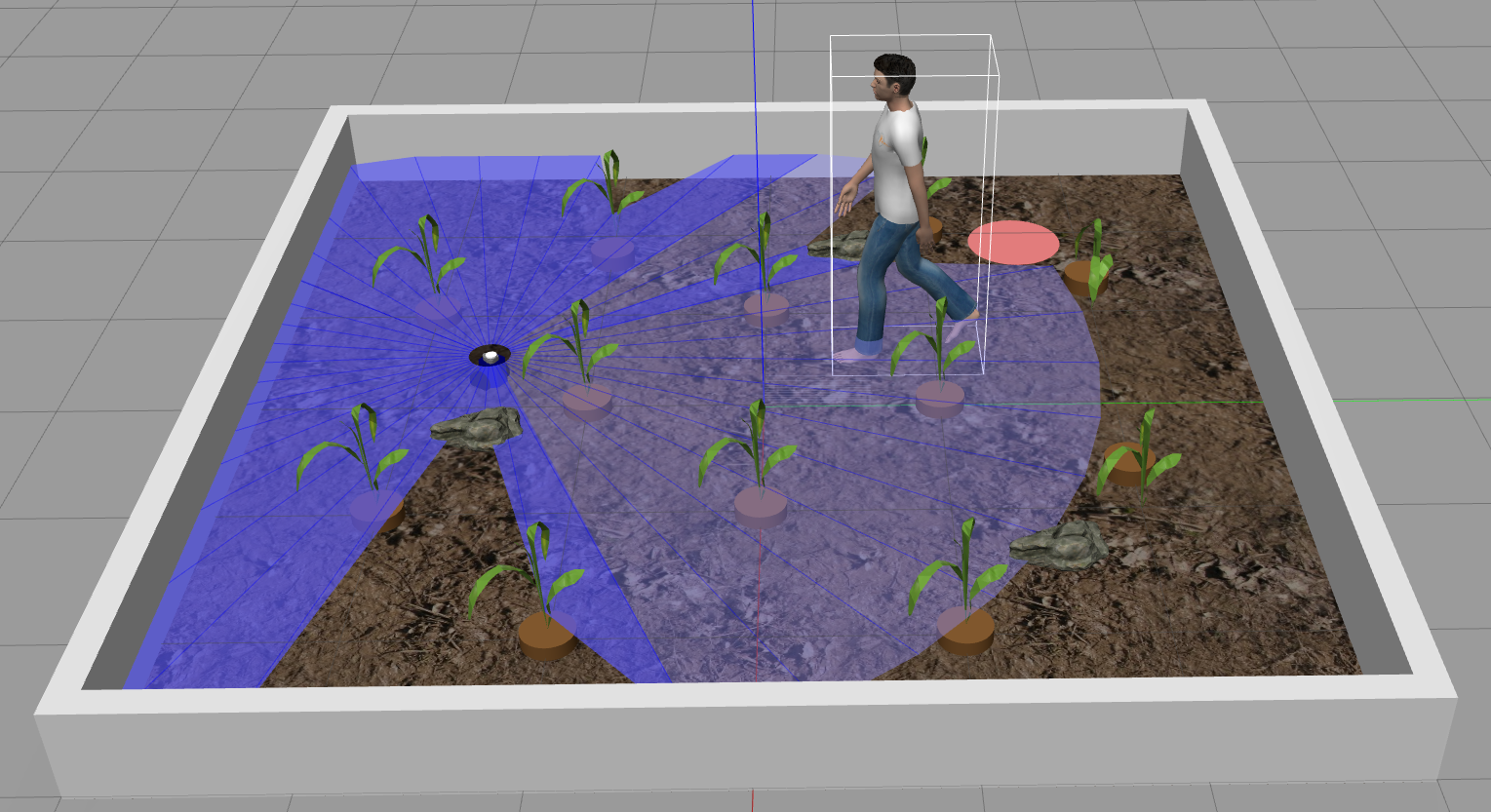}}
        \caption{Goal 3}
        \label{fig:Goal 3}
    \end{subfigure}
    % \hfill
    \begin{subfigure}[t]{0.24\textwidth}
        \centering
        \fbox{\includegraphics[width=\linewidth, height=4cm]{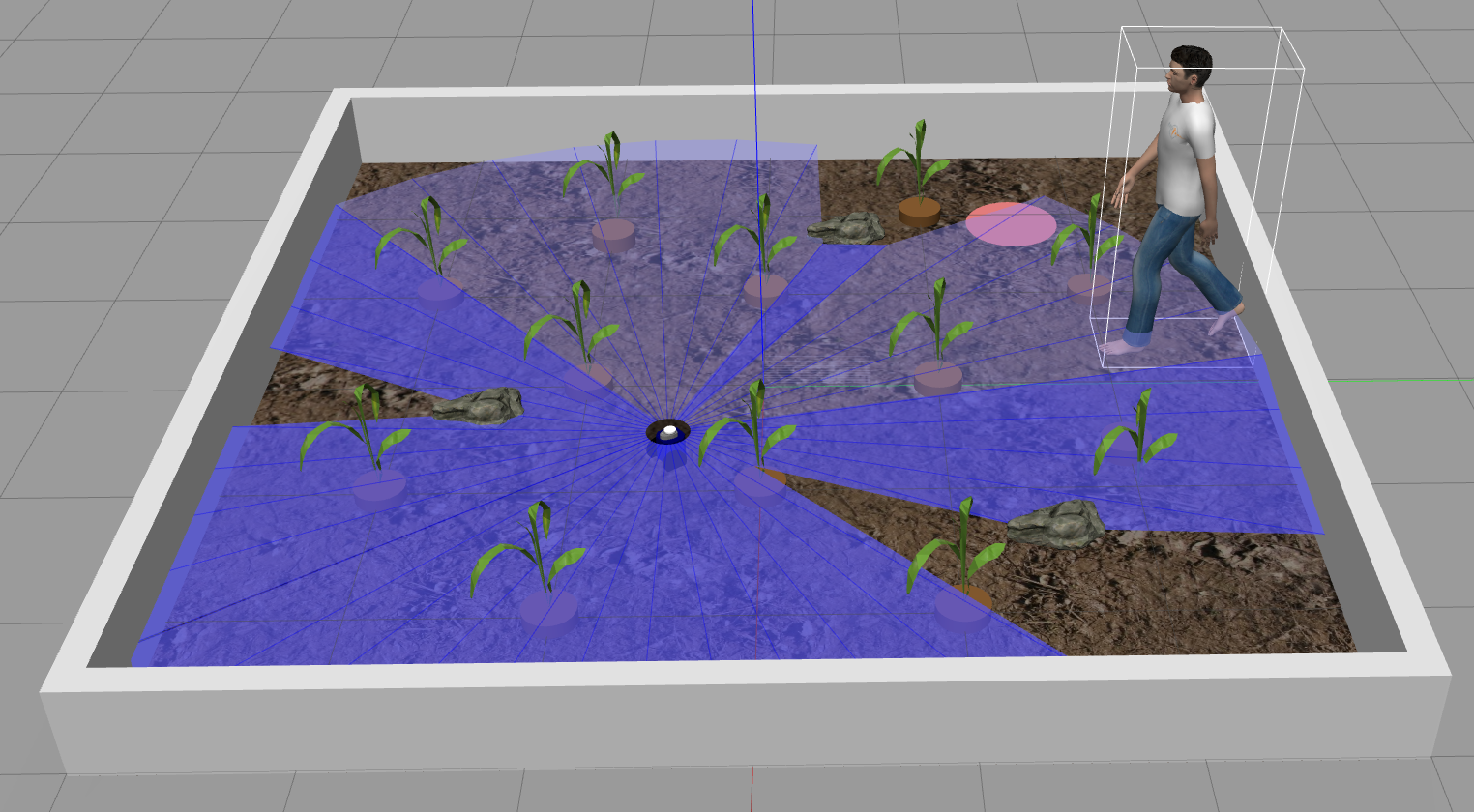}}
        \caption{Goal 4}
        \label{fig:Goal 4}
    \end{subfigure}
    \caption{Goals in static \& dynamic environem}
    \label{fig:goals-experiment}
\end{figure}

\subsection{Results and Analysis}

The performance of the static and dynamic agents is summarized in \cref{table:metrics-grouped}, which includes key metrics such as trailing 100 successes, trailing 100 rewards, and the time taken for training. Below is a detailed analysis of the results:

\begin{itemize}
    \item \textbf{Accuracy of Static Agent}
The success rate of static agents improves initially but decreases in later episodes. For example, \textbf{td3\_static\_700} achieves 85\% success, which drops to 76\% in \textbf{td3\_static\_900}. This decline is due to increased timeouts when goals are placed near crops, causing the agent to wait for the next goal to avoid crop damage.

\item \textbf{Reward Increase in Dynamic Agents}
Dynamic agents show a significant reward increase. For example, \textbf{td3\_dynamic\_500} has a reward of \textbf{-2595.982}, while \textbf{td3\_dynamic\_5000} achieves \textbf{1466.829}, a \textbf{156.5\% increase}. This improvement reflects the agent’s ability to navigate dynamic obstacles effectively.

\item \textbf{Pretrained Static Agent on Dynamic Environment}
When transferred to dynamic environments, pretrained static agents perform well. For instance, \textbf{td3\_dynamic\_5000} achieves a remarkable 95\% success rate. Notably, the unsuccessful outcomes are timeouts rather than collisions, ensuring no harm to the crops or the robot. Transfer learning allows the agent to adapt quickly to moving obstacles, significantly improving performance in fewer episodes.

    \item \textbf{Training Time}
Training time varies considerably between static and dynamic agents. The static agent requires upto 90 minutes to train upto 900 episodes while the dynamic agent takes upto 14 hours to train upto 5000 episodes!
\end{itemize}

\begin{table}[h]
    \captionsetup{justification=centering, font=small, labelsep=period, skip=0.5em} % Adjust caption properties
    \caption{Performance Metrics Grouped by Models and Agents}
    \label{table:metrics-grouped}
    \centering
    \small % Reduces font size to fit better in the column
    \renewcommand{\arraystretch}{1.2} % Adjust row height
    \begin{tabular}{|>{\centering\arraybackslash}p{2.5cm}|>{\centering\arraybackslash}p{3.5cm}|>{\centering\arraybackslash}p{2.5cm}|>{\centering\arraybackslash}p{2.5cm}|>{\centering\arraybackslash}p{2.5cm}|} \hline 
        \textbf{Model} & \textbf{Agent} & \textbf{Trailing 100 Successes} & \textbf{Trailing 100 Rewards} & \textbf{Timestamp} \\ \hline 
        \multirow{4}{*}{\textbf{Static}} 
            & td3\_static\_100 & 18\% & -3912.403 & 12:16 \\ \cline{2-5}
            & td3\_static\_300 & 72\% & -1075.812 & 35:57 \\ \cline{2-5}
            & td3\_static\_500 & 75\% & -101.037 & 56:34 \\ \cline{2-5}
            & td3\_static\_700 & 85\% & -1031.634 & 01:13:05 \\ \cline {2-5}
            & td3\_static\_900 & 76\% & -1423.000 & 01:31:02 \\ \hline 
        \multirow{4}{*}{\textbf{Dynamic}} 
            & td3\_dynamic\_500 & 19\% & -2595.982 & 01:35:17 \\ \cline{2-5}
            & td3\_dynamic\_1000 & 60\% & -1198.983 & 03:10:00 \\ \cline{2-5}
            & td3\_dynamic\_2000 & 80\% & 204.507 & 05:45:54 \\ \cline{2-5}
            & td3\_dynamic\_3000 & 86\% & 994.128 & 08:51:33 \\ \cline{2-5}
            & td3\_dynamic\_5000 & 95\% & 1466.829 & 14:03:08 \\ \hline
    \end{tabular}
\end{table}

\section{Discussion}
This research successfully achieved the following objectives:
\begin{enumerate}
    \item \textbf{Demonstrated the Efficiency of Deep Reinforcement Learning:} The study proved that deep reinforcement learning algorithms are more efficient than traditional methods by showcasing their adaptability and learning capabilities.

    \item \textbf{Enhanced Performance of Vanilla DQN:} The research improved the performance of the vanilla Deep Q-Network (DQN) by incorporating Double Q-Learning and Dueling Network architectures. This enhancement led to a 150\% faster convergence in the Double DQN (D3QN) compared to Double DQN (D2QN) and resulted in a 98\% reduction in training time within a complex 10x10 grid environment.

    \item \textbf{Comparison of TD3 and DDPG:} A comparative analysis of Twin Delayed DDPG (TD3) and Deep Deterministic Policy Gradient (DDPG) revealed an average stability improvement of 19.9\% in TD3. Additionally, TD3 outperformed DDPG in terms of mean path length and training time, making it the preferred choice for 3D simulations.

    \item \textbf{Development of a 3D Agricultural Environment:} The research team designed a 3D agricultural environment from scratch and implemented a TurtleBot3 agent using TD3. The agent demonstrated seamless obstacle avoidance in both static and dynamic scenarios and yielded a 95\% success rate in presence of moving obstacles.
\end{enumerate}

This study successfully extends prior work by incorporating dynamic obstacles and 3D simulations, elements suggested by \cite{zhao2023} but not addressed in their research . Our results demonstrate the robustness of DRL models in navigating complex, dynamic agricultural environments, highlighting the adaptability of these models.

\cite{lipeng2024} and \cite{gao2020} explored path planning in static environments using discrete action spaces . Gao’s incremental training from 2D to 3D for indoor robots and Lipeng’s DDQN with Prioritized Experience Replay (PER) are valuable for static environments but lack real-time adaptability. Our approach, focusing on continuous action spaces in dynamic agricultural fields, surpasses their methods by enabling smoother paths and more efficient decision-making in constantly changing settings.

\cite{raajan2020} and \cite{jiang2020} similarly employed discrete actions for collision avoidance in static environments. By introducing continuous action spaces, this study improves on their methods, yielding more adaptive solutions for real-time 3D scenarios, resulting in smoother navigation paths and enhanced decision-making in dynamic agricultural landscapes.

While \cite{haoliu2024} proposed TD3-DWA to optimize robotic navigation, focusing on velocity control in static settings, our work expands on this by implementing TD3 in dynamic environments . This shift allows for more adaptive, smoother paths, especially in agricultural scenarios, where obstacles are continuously changing.

Looking ahead, the next step involves implementing this solution on a real-life robot and testing it in actual agricultural fields. Additionally, this study opens the door for further enhancements, such as integrating the Dynamic Window Approach (DWA) with TD3 or experimenting with stochastic policy methods like SAC to improve decision-making under uncertainty.

\section{Declaration of competing interest}

The authors declare that they have no known competing financial
interests or personal relationships that could have appeared to
influence the work reported in this paper.

\section{Conclusion}
\label{conclusion}

The experiments conducted in this study demonstrate that learning-based algorithms, particularly Deep Q-Networks (DQN), significantly outperform deterministic primitive algorithms for Unmanned Ground Vehicle (UGV) path planning in precision agriculture. While DQN provided substantial benefits in adaptability and performance, the research identified several areas for further optimization. This led to the exploration of advanced versions such as Double DQN and Dueling Double DQN, which offer improvements in training stability and convergence speed.

However, recognizing the necessity for continuous action evaluation in complex agricultural environments, the study transitioned to a continuous 2D simulation. In this context, the Twin Delayed Deep Deterministic Policy Gradient (TD3) algorithm was found to outperform the Deep Deterministic Policy Gradient (DDPG), demonstrating enhanced performance in navigating the dynamic challenges of agricultural settings. Consequently, TD3 was selected as the optimal algorithm to enhance the agent's planning capabilities within a 3D environment utilizing Gazebo and ROS.

To effectively guide the agent’s decision-making, a custom reward structure and transfer-learning based approach was implemented, taking into account various environmental factors. This approach proved crucial for improving the agent's path planning efficiency. The results underline the potential of continuous deep reinforcement learning techniques in advancing UGV path planning for precision agriculture.

Future work will focus on transitioning from simulation to a live robot platform, enabling real-world testing in agricultural settings. This next phase will further validate the effectiveness of continuous deep reinforcement learning approaches in practical applications, emphasizing their role in enhancing the efficiency and reliability of UGV operations in precision agriculture.

\printcredits

\bibliographystyle{cas-model2-names}

\bibliography{main}

\end{document}